\PassOptionsToPackage{usenames, dvipsnames, table}{xcolor}
\documentclass[letterpaper, 11pt]{article}

\usepackage[margin=1in]{geometry}
\usepackage{setspace}\onehalfspace
\usepackage{microtype}

\usepackage{amsmath} \allowdisplaybreaks
\usepackage{amssymb}
\usepackage{amsthm} \theoremstyle{definition}

\newtheorem{proposition}{Proposition}

\usepackage{natbib}
\usepackage{bibentry}

\usepackage{booktabs}
\usepackage{graphicx}
\usepackage{appendix}

\usepackage[hyphens]{url}
\usepackage[bookmarks=false, hidelinks]{hyperref}

\usepackage{tikz}
\usetikzlibrary{positioning}
\usetikzlibrary{arrows, shapes}
\usetikzlibrary{decorations}
\usepackage{pgfplots}
\pgfplotsset{compat=1.18}

\usepackage{authblk}

\usepackage[shortlabels]{enumitem}
\hypersetup{colorlinks, citecolor=NavyBlue, linkcolor=NavyBlue, urlcolor=NavyBlue}
\usepackage[nameinlink, capitalise, noabbrev]{cleveref}
\hypersetup{colorlinks={false},linkcolor={blue},citecolor=green}

\crefname{equation}{}{}

\usepackage{subcaption}
\usepackage{multirow}
\usepackage{algorithm}
\usepackage{algpseudocode}

\usepackage{float}
\newfloat{algorithm}{t}{lop}

\algrenewcommand\algorithmicrequire{\textbf{Input:}}
\algrenewcommand\algorithmicensure{\textbf{Output:}}

\algnewcommand\algorithmicforeach{\textbf{for each}}
\algdef{S}[FOR]{ForEach}[1]{\algorithmicforeach\ #1\ \algorithmicdo}

\usepackage{wrapfig}

\newcommand{\email}[1]{{\href{mailto:#1}{\nolinkurl{#1}}}}

\usepackage{bm}

\newcommand{\ours}{NeuralSEP}
\newcommand{\bluenote}[1]{{\color{black}#1}}

\usepackage{etoolbox}
\makeatletter
\def\do#1{\@namedef{#1c}{\ensuremath{\mathcal{#1}}}}
\docsvlist{A,B,C,D,E,F,G,H,I,J,K,L,M,N,O,P,Q,R,S,T,U,V,W,X,Y,Z}
\makeatother

\renewcommand{\bar}[1]{\mkern 1.5mu\overline{\mkern-1.5mu#1\mkern-1.5mu}\mkern 1.5mu}
\renewcommand{\hat}{\widehat}
\renewcommand{\tilde}{\widetilde}

\DeclareMathOperator*{\minimize}{minimize}

\DeclareMathOperator*{\argmax}{\arg \max}
\DeclareMathOperator*{\argmin}{\arg \min}

\begin{document}

\title{A Neural Separation Algorithm for the Rounded Capacity Inequalities}

\date{}

\author[$\dag$]{Hyeonah Kim}
\author[$\dag$]{Jinkyoo Park}
\affil[$\dag$]{Department of Industrial and Systems Engineering, Korea Advanced Institute of Science and Technology (KAIST), 
\email{{hyeonah_kim, jinkyoo.park, chkwon}@kaist.ac.kr}}
\author[$\dag\ddag$]{Changhyun Kwon\thanks{Corresponding author.}}
\affil[$\ddag$]{Department of Industrial and Management Systems Engineering, University of South Florida}

\maketitle

\begin{abstract}
The cutting plane method is a key technique for successful branch-and-cut and branch-price-and-cut algorithms that find the exact optimal solutions for various vehicle routing problems (VRPs). Among various cuts, the rounded capacity inequalities (RCIs) are the most fundamental. To generate RCIs, we need to solve the separation problem, whose exact solution takes a long time to obtain; therefore, heuristic methods are widely used. We design a learning-based separation heuristic algorithm with graph coarsening that learns the solutions of the exact separation problem with a graph neural network (GNN), which is trained with small instances of 50 to 100 customers. We embed our separation algorithm within the cutting plane method to find a lower bound for the capacitated VRP (CVRP) with up to 1,000 customers. We compare the performance of our approach with CVRPSEP, a popular separation software package for various cuts used in solving VRPs. Our computational results show that our approach finds better lower bounds than CVRPSEP for large-scale problems with 400 or more customers, while CVRPSEP shows strong competency for problems with less than 400 customers.

\paragraph{Summary of Contribution:} 
We suggest a novel learning-based separation algorithm for RCIs arising in solving CVRPs. While some attempts have been made to learn to select cuts, our study is the first attempt to learn to generate cuts. In particular, we suggest a scalable model that leverages a message passing GNN with graph coarsening. The GNN with a sparse graph allows the trained model to solve problems of various sizes, while the graph coarsening reduces the size of graphs and enables learning representation from equivalent but non-isomorphic graphs. We prove that our model has a polynomial worst-case time complexity at the inference phase. Furthermore, we experimentally verify that our model has the transferability to solve problems sampled from out-of-distribution.
\end{abstract}

\section{Introduction} \label{sec:intro}
Cutting planes are the key components of many successful exact algorithmic frameworks for the integer programming (IP) formulations of various vehicle routing problems (VRPs): the capacitated VRP (CVRP), the VRP with time windows (VRPTW), the pickup-and-delivery problem (PDP), and their extensions.
When the classical branch-and-bound framework is combined with the cutting plane method, it forms the branch-and-cut (BC) algorithm \citep{laporte1985optimal};
when also combined with column-generation approaches,
it forms the branch-price-and-cut (BPC) algorithm \citep{fukasawa2006robust}.
The most successful exact algorithms are the BPC algorithms, which combine numerous techniques found in the last few decades, including effective cut generations, faster pricing algorithms, acceleration strategies for pricing, strong branching strategies, variable fixing, route enumeration, etc. 
See \citet{toth2014vehicle} and \citet{costa2019exact} for the descriptions of the various components in the BC and BPC algorithms.

As VRPs are $\mathcal{NP}$-hard,
it is inevitable that the exact algorithms have long computational times for large-scale problems.
Recently, various attempts have been made to apply advanced machine learning (ML) tools to improve and accelerate various components of the exact algorithms that were designed with the domain knowledge of an expert. 
Examples include \emph{column selection} \citep{morabit2021machine} and \emph{column generation} \citep{zhang2022learning} for VRPTW, and \emph{cut selection} \citep{tang2020reinforcement, paulus2022learning} and \emph{branching strategy} \citep{khalil2016learning} for general IP problems. 
They have demonstrated the potential of ML methods for accelerating the exact algorithms for VRPs.

In this paper, we develop an ML-based algorithm for another component of the exact algorithm that has not been considered in the literature: \emph{cut generation}.
Both BC and BPC algorithms rely on linear programming (LP) relaxations of the original IP problem and add inequality constraints iteratively.
Given a fractional solution of a relaxed LP problem, we need to find an inequality that \emph{separates} the fractional solution from the feasible region.
This problem is called the \emph{separation problem}, and such an inequality constraint, violated by the fractional solution but not by the feasible solutions of the original integer problem, is called a \emph{cut}.
These cuts, making the dual bounds tighter, are one of the key contributing factors to the success of BC and BCP algorithms.

Particularly, we focus on the separation problem for the \emph{rounded capacity inequalities} (RCIs) among many other cuts, such as framed capacity inequalities, strengthened comb inequalities, multi-star inequalities, subset row cuts, strengthened capacity cuts, and so on.
While utilizing various cuts, BC and BPC algorithms almost always begin by applying RCIs and then find other cuts if no more RCIs are identified. 
For this reason, an effective and efficient separation algorithm for RCIs plays a critical role in the exact algorithms.

The separation problem for RCIs is $\mathcal{NP}$-hard \citep{diarrassouba2017complexity}.
Although an exact method exists \citep{fukasawa2006robust}, heuristic separation algorithms are more popular; most notably, the heuristic algorithm by \citet{lysgaard2004new} and its open-source library CVRPSEP \citep{lysgaard2003cvrpsep}.
For problems with less than about 100 customer vertices, it is known that the bound found by CVRPSEP is almost the same as that found by the exact separation \citep{fukasawa2006robust, wenger2003generic}.
However, for larger problems, the performance of CVRPSEP for RCIs has hardly been evaluated since the exact separation requires an IP problem that takes a long time to solve.

For large-scale problems with {$400$} to $1,000$ customers, this paper shows that we can design a learning-based heuristic algorithm with better performance than CVRPSEP.
We design a novel neuralized separation algorithm, called \emph{\ours{}}, which learns the exact RCI separation algorithm through a graph neural network (GNN) and graph coarsening.
The proposed model finds effective cuts within a relatively short period of time {compared to the exact separation} for large-scale problems. 
To show the effectiveness of our approach, we embed \ours{} within the cutting plane method and obtain dual bounds, which will be compared with the dual bounds obtained by RCIs from CVRPSEP.
Our key contributions are as follows:
\begin{enumerate}
    \item We suggest a learning-based separation algorithm, the first of its kind to the best of our knowledge. Although some studies have already suggested learning-based cutting plane methods, they focus on learning to \emph{select} which cuts to add to the relaxed problem, and not to \emph{identify} the cuts by solving the separation problem.
    \item We suggest a scalable model leveraging GNN and graph coarsening. GNN with a sparse graph allows the trained model to solve problems of different sizes. Moreover, graph coarsening reduces the size of the graphs to handle large-sized graphs. Thus, \ours{} can efficiently solve the separation problem for RCIs in large-scale CVRPs.
    \item We prove that \ours{} has polynomial worst-case time complexity at the inference phase. 
    The RCI separation problem involves searching for subsets of vertices to find the subset that violates capacity constraints. \ours{} amortizes this computation by effectively learning representations during training; graph coarsening allows the model to learn graph representations from equivalent but non-isomorphic graphs. 
    \item \ours{} has the transferability to solve out-of-distribution problems---problems not sampled from the distribution used in the training phase.
    We experimentally verify the transferability of \ours{} by showing that the model trained with uniformly distributed demand can solve X-instances in CVRPLIB without additional training.
\end{enumerate}

\cref{sec:related_works} reviews the related works, and \cref{sec:preliminaries} introduces CVRP, RCIs, the cutting plane method, and existing separation methods for RCIs.
\cref{sec:method} explains the methodology of \ours{} with the training strategy in \cref{sec:training}.
Extensive computational experiments in \cref{sec:exp} verify the scalability, transferability, and effectiveness of \ours{}.
\cref{sec:conclusion} concludes this paper.

\section{Related Works} \label{sec:related_works}

This paper proposes a learning-based separation algorithm for RCIs.
As the separation problem is a combinatorial optimization (CO) problem, we introduce related works that apply learning-based methods, mainly with neural networks, to CO.
Solving CO problems with neural networks is not a new idea, as summarized in \citet{smith1999neural}.
However, it has gained more attention recently with the breakthrough of deep learning \citep{bengio2021machine,kotary2021end,bogyrbayeva2022learning}.

We review the learning-based CO algorithms in two categories: \emph{end-to-end} and \emph{hybrid learning} approaches.
The \emph{end-to-end approaches} train models to map the CO problems directly to their corresponding (optimal) solutions, while the \emph{hybrid learning approaches} use learning approaches to enhance existing exact/heuristic algorithms or solve subproblems arising within existing algorithms.
We conclude this section by introducing existing works related to RCIs briefly.

\subsection{End-to-End Machine Learning Approaches for CO}

For routing problems, learning-based approaches with sequential decision-making have been explored broadly \citep{vinyals2015pointer,bello2017neural,khalil2017learning,kool2018attention,park2021schedulenet,kim2022sym}.
They leverage recurrent neural networks \citep{bello2017neural}, Transformer \citep{kool2018attention,kim2022sym}, or graph neural networks \citep{khalil2017learning,park2021schedulenet} to learn the representations of CO problems.
Typically, they construct a route by appending a vertex to the current partial route one by one; since they choose which vertex to append based on the previous decisions, they are called auto-regressive models.

For CO problems, especially where decisions are made at the vertex level, some studies have suggested models that require fewer decision-making steps than auto-regressive models. 
\citet{schuetz2022combinatorial} proposed a one-shot prediction model that predicts soft (i.e., relaxed to continuous value) vertex assignments for the maximum-cut and maximal independent set (MIS) problems, then projects the soft assignments to binary values via rounding.
However, in general, the complicated constraints in CO problems are non-trivial in one-shot prediction methods with simple projection methods. 
\citet{ahn2020learning} proposed the learning-what-to-defer (LwD) method that iteratively decides on multiple decision variables at the same time by leveraging locally decomposable properties. LwD reduces the size of the problems by eliminating the variables decided in the previous iterations. Though LwD outperforms other end-to-end ML methods in MIS-related tasks, it requires a problem-specific transition function. 

\subsection{Hybrid Learning Approaches for CO}
\bluenote{ML can be utilized to address CO problems by incorporating learning-based methodologies with pre-existing algorithms.}
To solve routing problems, large neighborhood search methods \citep{shaw1997new,pisinger2010large} have been widely studied to improve their performances via neural networks.
\citet{chen2019learning} and \citet{lu2019learning} utilized deep learning methods to decide on which algorithm to use to destroy and repair the current solution at each step. 
\citet{hottung2020neural} employed the attention model of \citet{kool2018attention} to learn to repair the randomly broken partial solutions.
On the other hand, \citet{nair2020neural} and \citet{wu2021learning} focused on training the destroy operations via neural networks.

Some researchers show that deep learning can strengthen the exact algorithms to solve general IP problems:
for example, to decide on the primal heuristic strategy \citep{khalil2017primal,chmiela2021learning}, to set the branching score weight \citep{balcan2018learning}, and to select variables to branch \citep{gasse2019exact,gupta2020hybrid}.
A few studies enhanced the cutting plane method.
\citet{baltean2018selecting} introduced a neural network estimator that predicts the effectiveness of each cut and then selects the cuts to add based on this estimation.
\citet{tang2020reinforcement} suggested a neural network that selects a constraint to be applied within Gomory's cut procedure.
\citet{paulus2022learning} proposed the neuralized scorer that predicts the cost improvement of the relaxed problem with each cut.
These works have focused on the selection of cuts generated by the existing separation algorithm, while we generate new cuts directly to add to the relaxed problem.

Exact algorithms for VRPs have also been enhanced by deep learning. 
For the column generation methods as in the branch-and-price or branch-price-and-cut algorithms, learning methods for column selection \citep{morabit2021machine} and column generation \citep{zhang2022learning, morabit2022machine} for VRPTW have been studied.
Learning-based cutting plane enhancements specific to VRPs can hardly be found.

\subsection{Rounded Capacity Inequalities (RCIs)}

Introduced by \citet{laporte1983branch}, RCIs have played a prominent role in branch-and-cut algorithms.
\citet{augerat1995computational} suggested the first complete branch-and-cut algorithm for CVRP using RCIs, and
\citet{augerat1998separating} suggested the Tabu search algorithms, the first meta-heuristics for the RCI separation problems.
\bluenote{A branch-and-cut algorithm with heuristic separations of RCIs is proposed by \citet{ralphs2003capacitated}.}
\citet{lysgaard2004new} strengthened heuristics for RCIs and many other cuts and developed the CVRPSEP library \citep{lysgaard2003cvrpsep}.
Particularly, they generalized the `safe-to-shrink' condition for separation heuristics to reduce the graph size by utilizing the submodularity of the cost function of the separation problem.
Shrinking heuristics employ edge contraction that selects the edges and merges their endpoints into one vertex, which the graph coarsening procedures in \ours{} also utilize.
The graph coarsening in \ours{} differs from the shrinking heuristics in that it selects the edges based on the GNN predictions, whereas the shrinking heuristics select the edges that satisfy the safety condition---See \cref{sec:cvrpsep} for further discussions.
\citet{diarrassouba2017complexity} proved that the separation problem for RCIs {is} $\mathcal{NP}$-hard in general, and strongly $\mathcal{NP}$-hard when the demands are all the same.

\section{Problem Definition} \label{sec:preliminaries}

In this section, we formally introduce the definition of CVRP and RCIs and briefly illustrate the framework of the BC algorithm.
Then, we introduce the formulation of the exact separation method for RCIs, followed by a description of the heuristic separation method of CVRPSEP 
\citep{lysgaard2004new}.

\subsection{CVRP and RCIs} \label{sec:CVRP}

\bluenote{The capacitated vehicle routing problem, denoted as CVRP \citep{dantzig1959truck}, can be formalized as a tuple $(G, K, Q)$. Here, $G=(V, E)$ represents a complete undirected graph, where the vertex set $V$ is composed of vertices containing both the depot and customers, and the set $E$ is a set of the edges connecting pairs of vertices. We let $K$ denote the number of vehicles, each possessing a capacity constraint of $Q$.}
We let vertex 0 denote the depot, and $V_C = V \setminus \{0\}$ denote the set of customer vertices.
Each customer vertex $i$ has demand $d_i$ and each edge $(i, j)$ has non-negative cost $c_{ij}$. 
A route is defined as a set of edges that form a cycle {containing the depot}, and the cost of a route is the summation of all the costs of the edges that constitute the route.
\bluenote{
The goal of CVRP is to minimize the total cost of routes such that (i) every customer is assigned to only one vehicle for service, (ii) each route initiate and terminates at the depot, and (iii) the total demand of customers served by a vehicle does not exceed the capacity $Q$.}

Mathematically, CVRP can be formulated as an integer program (IP).
Following \citet{lysgaard2004new}, we define the edge variable $x_{ij}$ to represent the count of travels between vertices $i$ and $j$; note that $x_{ij} = x_{ji}$ since the edges are undirected.
\bluenote{For any customer subset $S \subseteq V_C$, we let $\delta(S)$ denote the set of \emph{crossing} edges whose endpoints belong to different sets, meaning one is in $S$, and the other is in $V \setminus S$.
Given a subset of edges $F \subseteq E$, we denote $x(F)$} as the sum of edge values in $F$, i.e., $\sum_{(i, j) \in F} x_{ij}$.
The \emph{two-index formulation} of CVRP is:
\begin{align}
    \minimize \quad & \sum_{(i, j) \in E} c_{ij} x_{ij} \label{cvrp_obj} \\
    \text{subject to} \quad & x(\delta ( \{i \})) = 2 && \forall i \in V_C \label{cvrp_c1} \\
     & x(\delta(S)) \geq 2 b(S) && \forall S \subseteq V_C \label{cvrp_c2} \\
     & x_{ij} \in \{0, 1\} && \forall 1 \leq i < j \leq |V| \label{cvrp_c3} \\
     & x_{0j} \in \{0, 1, 2\} && \forall j \in V_C, \label{cvrp_c4}
\end{align}
\bluenote{
where $b(S)$ denotes the minimum number of vehicles required to cover all demand in a customer set $S$.
The equality constraint presented in \Cref{cvrp_c1} ensures a requirement that every customer vertex has a degree of exactly 2, implying that each customer is visited precisely once by a vehicle.
As $b(S)$ is computed to serve customers in $S$ for the given capacity $Q$, the \textit{capacity constraint} in \cref{cvrp_c2} imposes that a route is connected to the depot by eliminating sub-tours. Lastly, the constraints detailed in \cref{cvrp_c3,cvrp_c4} correspond to the integer conditions for the decision variables.
Here, the edge variable connecting the customer $j$ and the depot, i.e., $x_{0j}$, has a value of 2 in cases where a vehicle exclusively serves a single customer. 
}

To solve CVRP, we first need to compute $b(S)$ by solving the bin packing problem whose bin capacity and sizes of each item are $Q$ and $d_i$, respectively.
It is well-known that calculating $b(S)$ is strongly $\mathcal{NP}$-hard.
Fortunately, the formulation is still valid---the set of feasible integer solutions \bluenote{remains the same}---even if $b(S)$ is replaced with $k(S) = \lceil \sum_{i \in S} d_i/Q \rceil$, which is equal to or smaller than $b(S)$.
The constraints in \cref{cvrp_c2} with $k(S)$ are called the \emph{rounded capacity inequalities} (RCIs).

\subsection{The Branch-and-Cut Algorithm} \label{sec:cutting_plane}
Even if the two-index formulation with RCIs is well defined, the RCIs cannot be used directly in practice.
The main reason is that the number of RCIs grows exponentially with respect to $|V|$. 
The branch-and-cut algorithm handles such complexity in an iterative manner via the cutting plane method.

\bluenote{
The cutting plane method starts with relaxing a given problem by disregarding the hard-to-hand constraints. In CVRP, the RCIs and the integer conditions associated with decision variables are usually relaxed \citep{ralphs2003capacitated,lysgaard2004new}. Then, the cutting plane method solves the relaxed problems and finds the violated constraints among disregarded constraints. The violated constraints are called the cutting planes or simply cuts. 
To yield a new relaxed problem for the subsequent iteration, the cutting planes are added to the relaxed problem. This iterative process continues until the solution of the relaxed problem, referred to as the relaxed solution, satisfies the feasibility criteria of the original problem.}
If the cutting plane method cannot find an integer solution, we start a branch-and-bound scheme to complete the branch-and-cut algorithm.

\bluenote{
In the context of the two-index formulation of CVRP, the relaxed problem is formulated with linear programming (LP) as follows:}
\begin{align}
    \minimize \quad & \sum_{(i, j) \in E} c_{ij} x_{ij} \label{relaxed_obj} \\
    \text{subject to} \quad & x(\delta ( \{i \})) = 2 && \forall i \in V_C \label{relaxed_c1} \\
     & 0 \leq x_{ij} \leq 1 && \forall 1 \leq i < j \leq |V| \label{relaxed_c2} \\
     & 0 \leq x_{0j} \leq 2 && \forall j \in V_C.\label{relaxed_c3}
\end{align}

\begin{figure}
  \centering
  \includegraphics[width=0.95\textwidth]{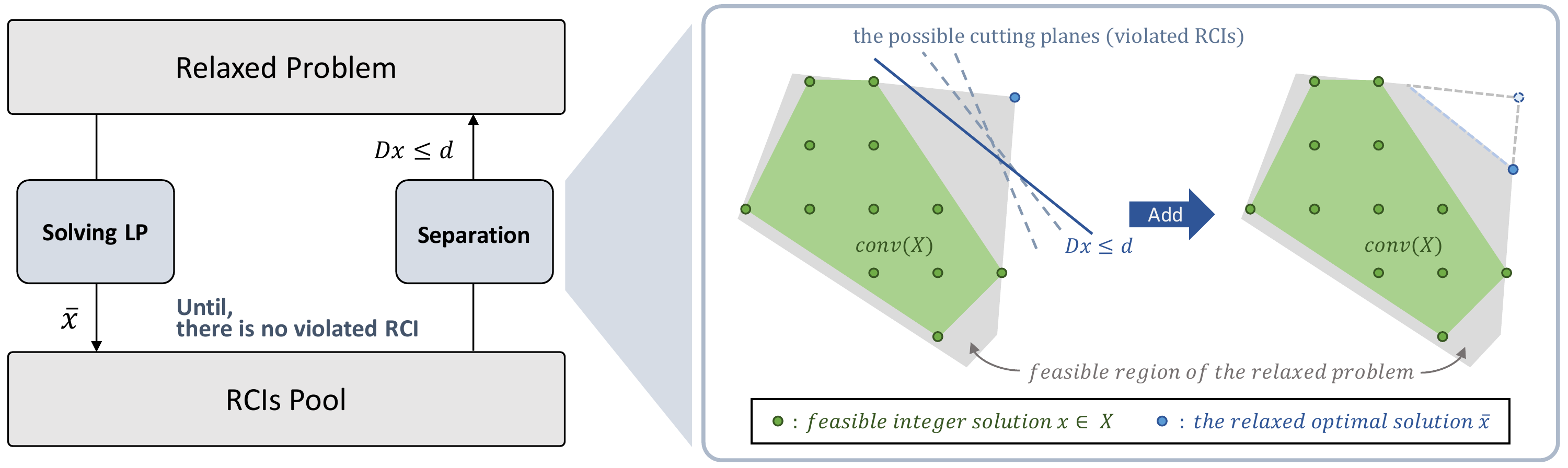}
  \caption{Illustration of the iterative separation procedure in cutting plane method}
  \label{fig:cutting_plane_method}
\end{figure}

\bluenote{The RCIs are generated and added to the above LP problems, \crefrange{relaxed_obj}{relaxed_c3}, iteratively. Since the RCIs and integer conditions are dropped, the relaxed solution (denoted as $\bar{x}$ in \cref{fig:cutting_plane_method}) is likely infeasible for the original CVRP problem \crefrange{cvrp_obj}{cvrp_c4}. Therefore, a separation algorithm refines the feasible region of the relaxed problem by adding an RCI, $Dx \leq d$, which is valid to the original integer solutions, but not valid to the current relaxed solution, i.e., $D\bar{x} > d$.
The RCI \emph{separates} the relaxed solution from the original feasible solutions.}
The cutting plane method repeats this procedure until the separation algorithm fails to find any more RCIs, as demonstrated in \cref{fig:cutting_plane_method}.

\subsection{Exact Separation Algorithm for RCIs} \label{sec:exact_separation}
The exact separation algorithm \citep{fukasawa2006robust} is designed to produce the most violated RCIs.
\bluenote{The exact separation algorithm refines the feasible region of the relaxed problem by adding the most violated RCIs, while heuristic algorithms focus on finding possible RCIs rapidly.}

\bluenote{The separation problem is mathematically formulated as a mixed-integer program (MIP).} We start by defining the notations of the exact formulation.
\bluenote{For a given relaxed solution $\bar{x}$, we define a support graph as $\bar{G}=(V, \bar{E})$ where $\bar{E} = \{(i, j) \in E:\bar{x}_{ij} > 0\}$.
In addition, we introduce binary variables $y_i$ for all $i \in V$, and continuous variables $w_{ij}$ for all $(i, j) \in \bar{E}$.
We let $y_i = 1$ when the vertex $i$ belongs to the set $S$, and $w_{ij} = 1$ when the edge is in the boundary of set $S$, i.e., $(i, j) \in \delta(S)$.
For every $M \in \{0, \ldots, \lceil \sum_{i\in V_C} d_i / Q \rceil - 1 \}$, we get the minimum weight of crossing edges, i.e., $z(M)$,} by solving the following problem:
\begin{align}
    z(M)\ = \
     \min \quad & \sum_{(i, j) \in \bar{E}} \bar{x}_{ij} w_{ij} \label{exact_rci_obj} \\
    \text{s.t.} \quad & w_{ij} \geq y_i - y_j && \forall (i, j) \in \bar{E} \label{exact_rci_c1} \\
     & w_{ij} \geq y_j - y_i && \forall (i, j) \in \bar{E} \label{exact_rci_c2} \\
     & \sum_{i \in V_C} d_i y_i \geq (M \cdot Q) + 1 \label{exact_rci_c3} \\
     & y_0 = 0 \label{exact_rci_c4} \\
     & y_i \in \{0, 1\} && \forall i \in V_C \label{exact_rci_c5} \\
     & w_{ij} \geq 0 && \forall (i, j) \in \bar{E} \label{exact_rci_c6}
\end{align}
\bluenote{
The constraints presented in \cref{exact_rci_c1,exact_rci_c2} indicate that the variable $w_{ij}$ equals 1 if the endpoints belong to distinct sets, i.e., $y_i \neq y_j$, which is the definition of $w_{ij}$.
\cref{exact_rci_c3} means that the total demand of $S$ exceeds the total capacity, i.e., $M\cdot Q$, violating the capacity constraints. Consequently, the set $S$ necessitates allocating a minimum of $M+1$ vehicles to serve the demands within $S$. In other words, the subset $S$ requires traversal more than $2(M+1)$ times, so we can find the RCI if $z(M) < 2(M+1)$. The violated RCI has the form of $x(\delta(S^*)) \geq 2 \lceil\sum_{i\in S^*}d_i/Q \rceil$, where $S^*=\{i \in V_C : y_i^* = 1\}$.}

For each $M$, the separation problem needs to find an optimal subset $S^*$ that has the minimum crossing edge weight, so it is a CO problem.
The separation problem for RCIs is $\mathcal{NP}$-hard and strongly $\mathcal{NP}$-hard in the case of unit demand  \citep{diarrassouba2017complexity}.

\subsection{Heuristic Separation Algorithms for RCIs} \label{sec:cvrpsep}
Designing effective heuristics has been the focus of the separation problems for RCIs due to its intractability.
CVRPSEP, implemented based on \citet{lysgaard2004new}, is a widely employed separation algorithm package not only for RCIs but also framed capacity inequalities, strengthened comb inequalities, multi-star inequalities, and hypotour inequalities.
In this section, we review the heuristic algorithms for RCIs in CVRPSEP, which employs four different heuristic algorithms in order.

The first heuristic is the connected components heuristic.
Early versions of the connected components heuristic were proposed independently in \citet{ralphs1995parallel} and \citet{augerat1998separating}.
In CVRPSEP, the connected components heuristic obtains $\bar{G}_C = (V_C, \bar{E}_C)$ by removing the depot from the given support graph $\bar{G}$.
Then, it finds the connected components $S_1, \ldots, S_p$ of $\bar{G}_C$ and checks whether the RCI corresponding to the component is violated.
Specifically, for $i=1, \ldots, p$, the algorithm checks whether the connected component $S_i$ and its complement $V_C \setminus S_i$ violate the RCI.
If all $S_i$ and $V_C \setminus S_i$ satisfy the RCIs, {the union of those connected components that are not connected to the depot in the support graph} is checked.

When the connected components heuristic fails, CVRPSEP makes the support graph \textit{shrink}.
It chooses a subset of customers $S$ and shrinks it to a vertex $s$, called a super-vertex, whose demand is set to $\sum_{i\in S}d_i$, and the weight of the edge $(s, j)$ in the shrunk support graph is set to $\sum_{i\in S} \bar{x}_{ij}$ for each $j \in V_C$.
Edge contraction is employed to obtain the shrunk graph, while the subset $S$ should satisfy certain `safety' conditions dependent on the current LP solution $\bar{x}$ to find a cut.
On the other hand, in this study, we suggest to coarsen the graph based on the GNN prediction so that the edges are selected according to the probabilities that the endpoints belong to the same sets.
For the shrunk support graph, additional three heuristics are used: fractional capacity inequalities, greedy construction heuristic, and removal heuristic;
See \citet{lysgaard2004new} for details.

\section{The Neural Separation Algorithm for RCIs} \label{sec:method}

In this section, we propose an ML-based separation algorithm for RCIs, \ours{}, and explain how \ours{} finds the subset corresponding to the violated RCIs for a given support graph $\bar{G}$.
We employ graph neural networks (GNN) with a message passing scheme \citep{gilmer2017neural}, which are able to extract embeddings by utilizing the graph structures.
The model is trained to imitate the exact separation algorithm in \cref{sec:exact_separation}.

\begin{figure}
  \centering
  \includegraphics[width=0.99\textwidth]{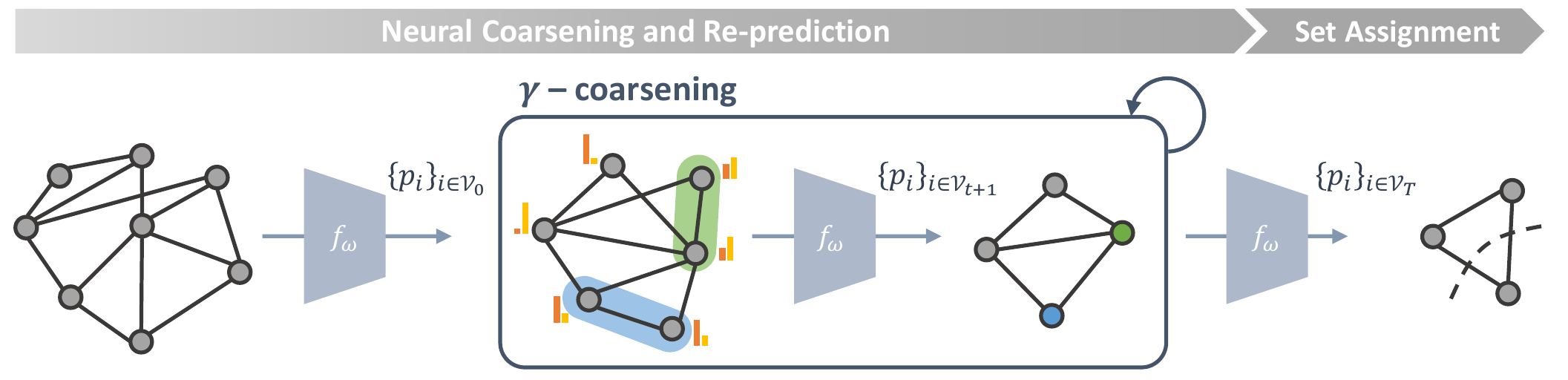}
  \caption{The overall coarsening procedure of \ours{}: It iteratively predicts vertex selection probabilities and coarsens the given graph. \ours{} decides the set assignment on the coarsest graph, then maps the assignment to the original graph.}
  \label{fig:coarsening}
\end{figure}

\cref{fig:coarsening} illustrates the overall process of \ours{}.
The model predicts a vertex selection probability $p_i$ that the vertex is in subset $S$
(\cref{sec:gnn} and \cref{sec:graph_embedding}); it can be interpreted as the continuous relaxed value of the decision variable $y_i$.
Then, the given graph is coarsened to a smaller graph depending on $p_i$, and the vertex selection probabilities are re-computed with the coarser graph (\cref{sec:coarsening_with_gnn}).
The coarsening process is repeated until the coarse graph has the depot and two other vertices only, or there {are} no edges to contract.
Vertices in the coarsest graph are assigned to $S$ or the complement of $S$, and lifted to the original graph (\cref{sec:set_assignment}).

\subsection{Message Passing Graph Neural Networks} \label{sec:gnn}
{
In this section, we provide a brief exploration of graph neural networks (GNN). We employ GNN with a \emph{message passing} scheme \citep{gilmer2017neural}, since the RCI separation problem is naturally defined on a graph.
For given graph $\mathcal{G}=(\mathcal{V}, \mathcal{E})$, assume that each vertex $i \in \mathcal{V}$ holds a vertex feature $h_i$ and each edge $(i, j) \in \mathcal{E}$ holds an edge feature $h_{ij}$.
GNN iteratively updates the features using the edge update function (i.e., message function) and vertex update function (i.e., message aggregation).

\paragraph{Edge update.} The edge feature $h^{(\tau)}_{ij}$ at iteration $\tau$ is updated as follows:
\begin{equation} \label{eq:gnn_e}
    h^{(\tau)}_{ij} = f_e \left(\left[h^{(\tau-1)}_{i}, h^{(\tau-1)}_{j}, h^{(\tau-1)}_{ij}\right]; \theta_e \right), \quad \forall (i,j) \in \mathcal{E}
\end{equation}
where $f_e$ is the edge update function, which is a Multi-layer Perceptron (MLP) parameterized with $\theta_e$, and $h^{(\tau-1)}_{i}$ and $h^{(\tau-1)}_{ij}$ are the vertex embedding and the edge embedding at the $\tau - 1$ iteration. The updated edge embedding is regarded as a message.

\paragraph{Vertex update.} Vertex $i$ receives messages from its neighbors (i.e., the vertices connected to $i$) $j \in \mathcal{N}(i)$, where $\mathcal{N}(i) = \{j \in \mathcal{V} : (i, j) \in \mathcal{E}\}$.
Each vertex gathers and aggregates the messages, and updates its vertex feature as follows:
\begin{equation} \label{eq:gnn_n}
    h^{(\tau)}_{i} = f_v \left(\left[h^{(\tau-1)}_{i}, \textsf{AGG}_{j \in \mathcal{N}(i)} \left(h^{(\tau-1)}_{ij} \right)\right]; \theta_v \right), \quad \forall i \in \mathcal{V}
\end{equation}
where $f_v$ is the vertex update function, which is an MLP parameterized with $\theta_v$, and $\textsf{AGG}$ is a differentiable and permutation invariant aggregate function (e.g., sum, mean, or max).
}

\subsection{Graph Embedding with GNN} \label{sec:graph_embedding}
At each separation step, \ours{} takes the current relaxed solution $\bar{x}$ and its support graph $\bar{G}$ as an input.
In the support graph, vertices $i$ and $j$ are connected only if $\bar{x}_{ij}$ is greater than $0$.
Since the support graph can be disconnected (i.e., there may exist connected components not containing the depot), we added edges between the depot and customers with $0$ values.
Thus, the support graph is connected while maintaining its sparsity.
\bluenote{It is noteworthy that if vertices are disconnected, messages are not exchanged; since the message corresponds to the edge embedding in GNN. To prevent this, we define the augmented support graph} as $\bar{G}'=(V, \bar{E}')$, where $\bar{E}' = \{(i, j) \in E:\bar{x}_{ij} > 0 \mbox{ or } i = 0\}$.
\bluenote{
To model the separation problem \crefrange{exact_rci_obj}{exact_rci_c6} within the graph framework, we define the vertex and edge features for $\bar{G}'$ using coefficients in the problem.
}

\paragraph{Vertex Feature.} 
\bluenote{
We utilize the coefficient $d$, and RHS $M$ and $Q$ in \cref{exact_rci_c3}, which is highly related to the vertex decision variable $y$.
The constraint can be reformulated as $\sum_{i \in V_C}{d_i} y_i / Q > M$. 
Here, the coefficient ${d_i}/{Q}$ inherently falls within the range of $[0, 1]$ due to its definition. 
To make the features have similar ranges, we normalize $M$ by the number of vehicles $K$.
In summary,} the vertex feature for vertex $i \in V$ is defined as follows:
\begin{equation}
    s_i = \left(\frac{d_i}{Q}, \frac{M}{K}\right)
\end{equation}

\bluenote{
\paragraph{Edge Feature.} We utilize the cost coefficient $\bar{x}$ in \cref{exact_rci_obj}, which is associated with the edge decision variable $w$. 
It is noteworthy that $\bar{x}_{ij}$ takes a value of $0$ for the additional edges, i.e., $(i, j) \in \bar{E}' \setminus \bar{E}$, and each $\bar{x}_{ij}$ falls within the range of $[0, 2]$. Consequently, the edge feature for $(i, j) \in \bar{E}'$ is defined as follows:}
\begin{equation}
    s_{ij} = \left(\bar{x}_{ij} \right)
\end{equation}

\paragraph{Computing the Vertex Selection Probability.}
\ours{} first encodes vertex feature $s_i$ and edge feature $s_{ij}$ to get the initial embedding:
\begin{align}
    h_{ij} = g_e(s_{ij};\phi_e) & \quad \forall (i, j) \in \bar{E}' \\
    h_i = g_v(s_i;\phi_v) & \quad \forall i \in V, 
\end{align}
where $g_v$ and $g_e$ are simple MLPs parameterized with $\phi_e$ and $\phi_v$, respectively.

\ours{} applies GNN parameterized with $\theta_v$ and $\theta_e$ to produce the final embedding $\Vec{H}'_v = \{ h'_i\}_{i \in V}$ as follows:
\begin{equation} \label{eq:gnn}
    \Vec{H}'_v = \textsf{GNN} \left( \Vec{H}_v, \Vec{H}_e; \theta_v, \theta_e \right),
\end{equation}
where $\Vec{H}_v$ and $\Vec{H}_e$ denote the vertex and edge initial embedding vectors (i.e., $\Vec{H}_v = \{h_i\}_{i \in V}$, and $\Vec{H}_v = \{h_{ij}\}_{(i, j) \in \bar{E}'}$).

Using the final embedding $h'_i$ for vertex $i$, \ours{} computes the probability that $i$ is included in subset $S$ as follows:
\begin{equation} \label{eq:gnn_p}
    p_{i} = \pi \left( h'_{i}; \theta_p \right), \quad \forall i \in V,
\end{equation}
where $\pi$ is an MLP with $\theta_p$.
\bluenote{We use a sigmoid activation at the final layer in $\pi$ to ensure the $p_{i}$ in $[0, 1]$.}
Note that the probability for each vertex is calculated independently.
Another way of interpreting $p_i$ is to regard it as a continuous relaxed value of the integer variable $y_i$ in the exact separation problem in \cref{sec:exact_separation}.
We can denote the entire process described above as $f_\omega(\bar{G}')$, where the parameter $\omega = (\phi_e, \phi_v, \theta_e, \theta_v, \theta_p)$ is learnable.

\subsection{Graph Coarsening with Re-prediction} \label{sec:coarsening_with_gnn}
Inspired by the shrinking heuristics \citep{augerat1998separating,ralphs2003capacitated,lysgaard2004new}, \bluenote{we utilize a graph coarsening procedure to discretize the predicted probabilities. We can identify the subset $S$ by iteratively coarsening graphs and re-predicting probabilities on the coarsened graph, we can identify the subset $S$.

The graph coarsening procedure transforms an initial graph $\mathcal{G}_0$ into a sequence of downsized graphs denoted as $\mathcal{G}_1, \mathcal{G}_2, \ldots, \mathcal{G}_T$, with the property that the number of vertices decreases over each iteration, i.e., $|\mathcal{V}_0| > |\mathcal{V}_1| > \cdots > |\mathcal{V}_T|$, where $T$ represents the number of iterations. In our context, the initial graph $\mathcal{G}_0$ corresponds to the augmented support graph $\bar{G}'$.
}

\begin{algorithm}
\caption{$\gamma$-coarsening}
\label{algo:coarsening}
\small
\begin{algorithmic}[1]
\Require A graph $\mathcal{G}_t=(\mathcal{V}_t, \mathcal{E}_t)$, contraction probability matrix $\bm{q}$, coarsening ratio $\gamma$
\Ensure The coarser graph $\mathcal{G}_{t+1} = (\mathcal{V}_{t+1}, \mathcal{E}_{t+1})$
    \State Initialize $\mathcal{G}_{t+1} \gets \mathcal{G}_t$
    \While{$|\mathcal{V}_{t+1}| > \lfloor \gamma \cdot |\mathcal{V}_t| \rfloor$}
        \If{$q_{ij} = 0$ for all $(i, j) \in \mathcal{E}_{t+1}$}
            \State Terminate the while-loop
        \Else
        \State $(u', v') \gets \argmax_{(i, j)\in \mathcal{E}_{t+1}} q_{ij}$
        \Comment{Select an edge to contract based on \cref{eq:edge_prob}}
        \State $\mathcal{N}(u') \gets \{v \mid (u', v) \in \mathcal{E}_{t+1} \}$
        \Comment{Find the neighbor vertices}
        \State $\delta(u') \gets \{(u', v) \in \mathcal{E}_{t+1} \mid v \in \mathcal{V}_{t+1} \}$
        \Comment{Find the connected edge set}
        \State $\mathcal{E}_{t+1} \gets \mathcal{E}_{t+1} \cup \{(v, v') \mid v\in \mathcal{N}(u)\}$ with edge feature $s_{vv'} = s_{vu'}, \forall v \in \mathcal{N}(u')$
        \Comment{Create new edges}
        \State $d_{v'} \gets d_{u'} + d_{v'}$
        \Comment{Merge vertex $u'$ into $v'$}
        \State $\mathcal{G}_{t+1} \gets \left( \mathcal{V}_{t+1} \setminus \{u'\}, \mathcal{E}_{t+1} \setminus \delta(u') \right)$
        \State Combine the parallel edges with weight summation to simplify $\mathcal{G}_{t+1}$
        \EndIf
    \EndWhile
    \\
    \Return $\mathcal{G}_{t+1}$
\end{algorithmic}
\end{algorithm}

\bluenote{
In this work, we coarsen graphs by merging two vertices $(u, v)$ into a single vertex denoted as $v'$; this is also called contraction. 
We set the weight of $v'$ as the sum of weights of $u$ and $v$. And if there are parallel edges, we combine them into a single edge with their values being summed.}
The graph is coarsened with ratio $\gamma < 1$, the $t$-th coarse graph $\mathcal{G}_t$ has $|\mathcal{V}_t| = \lfloor \gamma^{t-1} |\mathcal{V}_0| \rfloor$ vertices.
The set of pairs of vertices to contract is chosen based on the probability that both are in the same group ($S_t$ or $\mathcal{V}_t \setminus S_t$).
As the depot is excluded from set $S_t$ in the exact separation problem, we set the contraction probabilities of the depot-connected edges as $0$. 
The contraction probability $q_{ij}$ is defined as the probability of being excluded in the crossing edge set (i.e., the probability of connecting vertices in the same set) as follows:
\begin{equation} \label{eq:edge_prob}
    q_{ij} = \begin{cases}
        p_i p_j + (1-p_i)(1-p_j) & \mbox{if } i, j \neq 0 \\
        0 & \mbox{otherwise.}
    \end{cases}
\end{equation}
The pseudo-code of $\gamma$-coarsening is provided in \cref{algo:coarsening}.
With the coarser graph $\mathcal{G}_{t+1}$, we re-compute the vertex selection probability $p_i$ via $f_\omega(\mathcal{G}_{t+1})$.
We repeat $\gamma$-coarsening until there are three vertices (i.e., a depot and two aggregated vertices) remaining, or there are no pair of vertices to contract (i.e., the contraction probabilities $q_{ij}$ are all $0$).
\bluenote{Given that the graph is coarsened by a factor of $\gamma$ at each iteration, the number of predictions is $\mathcal{O}(\log |V|)$.}

The main difference from the shrinking heuristics in \citet{lysgaard2004new} is that we coarsen graphs based on the GNN predictions, not the relaxed solution values.
\ours{} iteratively coarsens the graph based on the GNN prediction and re-predicts the vertex selection probability based on the coarsened graphs.
\bluenote{Thus, we use the graph coarsening scheme to de-randomize the continuous prediction of GNN to identify a subset.}
We show that the graph coarsening preserves the properties of the separation problem when the error of GNN is bounded:
\begin{proposition} \label{prop:error}
    For any given $t$, if the prediction error of $f_\omega(\mathcal{G}_t)$ is bounded by $1/2$, then the graph coarsening process preserves the crossing edge weights and the total demand of the selected vertices.
\end{proposition}
We provide all proofs in the appendix.
The preservation means that the coarser graphs are equivalent to the original graphs when solving the separation problem.
However, we modify the structure of the graph by merging vertices and removing edges in the coarsening procedure; thus, the coarser graph has a different structure from the original graph, i.e., graphs are non-isomorphic.

\subsection{Set Assignment and Graph Uncoarsening} \label{sec:set_assignment}
After $T$ steps of the coarsening operation, the coarsest graph $\mathcal{G}_T=(\mathcal{V}_T, \mathcal{E}_T)$ that has the depot and at least two vertices (i.e., $|\mathcal{V}_T| \geq 3$) is obtained.
The final probabilities are computed via $f_\omega(\mathcal{G}_T)$; we then apply the following simple projection rules to map the probabilities $\{p_i\}_{i \in \mathcal{E}_T}$ back to integer variables $\{ y_i \}_{i \in \mathcal{E}_T}$:
\begin{itemize}
    \item[] \textbf{Step 1.} Set $y_i = 1$ if $p_i > 1/2$.
    \item[] \textbf{Step 2.} If there are no vertices with $y_i = 1$, set $y_j = 1$, where $j = \argmax_{i \in \mathcal{V}_T} \{ y_i\}$.
\end{itemize}
By this set assignment process, the vertices of $\mathcal{G}_T$ are bi-partitioned to set $S_T = \{i \in \mathcal{V}_T : y_i = 1 \}$ and its complement $\mathcal{V}_T \setminus S_T$.

\begin{figure}
  \begin{center}
    \subfloat[The final graph $\mathcal{G}_T$]{
    \includegraphics[width=0.3\linewidth]{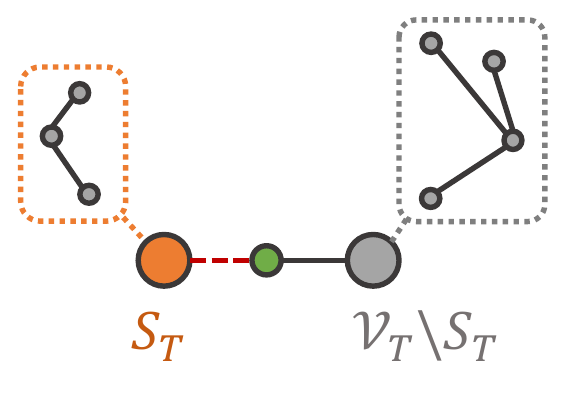}
    \label{fig:sT_sep}}
    \hfil
    \subfloat[The initial graph $\mathcal{G}_0$]{
        \includegraphics[width=0.3\linewidth]{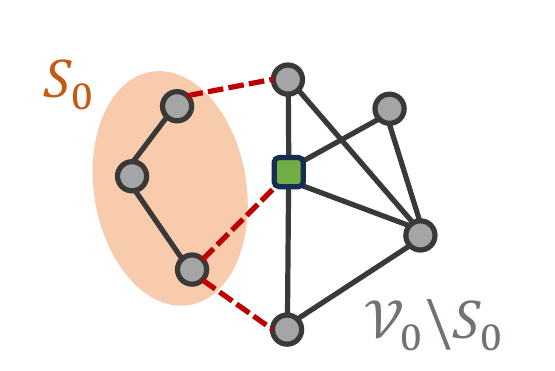}
    \label{fig:s0_sep}}
  \end{center}
  \caption{An example of the set assignment and uncoarsened results. The vertices are assigned to a set and its complement, and these set assignments are directly mapped to the initial graph $\mathcal{G}_0$. Note that the square vertex represents the depot.}
  \label{fig:uncoarsening}
\end{figure}

After the set assignment, inverse coarsening operations are conducted to lift the coarsest graph $\mathcal{G}_T$ back to its original graph, as shown in \cref{fig:uncoarsening}.
The super-vertices of the coarser graph $\mathcal{G}_t$ are composed of a distinct subset of vertices of the finer graph $\mathcal{G}_{t-1}$.
As the merged vertices information is tracked during the coarsening process, the assignment of the super-vertices in $\mathcal{G}_T$ can be mapped to the assignment of the vertices in $\mathcal{G}_0$ directly.
Since the weights are aggregated in the coarsening phase, the sum of the crossing edge weights and vertex weights are preserved in the lifting operation; that is, $\bar{x}(\delta(S_{T})) = \bar{x}(\delta(S_{0}))$ and $ \sum_{i\in S_{T}}d_i = \sum_{i\in S_{0}}d_i$, where $\bar{x}(F) = \sum_{(i, j) \in F} \bar{x}_{ij}$ for any edge set $F$.

Similar to the exact RCI separation, \bluenote{given $M = 0, \ldots, \lceil \sum_{i\in V_C} d_i / Q \rceil - 1$,} we find a subset $S_0(M)$ and examine whether its RCI is violated.
If $\bar{x}(\delta(S_0(M)) < 2 \lceil \sum_{i \in S_0(M)} d_i/Q \rceil$, then the corresponding RCI is violated; therefore, we add it to the current relaxed problem.
This examination process allows \ours{} to be free from the safety conditions in the coarsening procedure.

We conclude this section by showing that the worst-case time complexity of \ours{} is polynomial for inference:

\begin{proposition} \label{prop:complexity}
    The forward propagation of \ours{} has $\mathcal{O}(|\mathcal{V}| |\mathcal{E}| \log |\mathcal{V}|)$ worst-case time complexity.
\end{proposition}
Coarsening involves $T$ iterations of feature embedding and graph coarsening. 
The termination condition for the coarsening process is either 1) when the number of vertices reaches three or 2) when every edge has zero contraction probability (i.e., early termination). 
At each iteration $t$, the number of vertices is bounded by $\lfloor \gamma^{t}|\mathcal{V}_{t-1}| \rfloor$, where $\gamma$ represents the coarsening ratio. Accordingly, the \bluenote{maximum number of iterations is limited to} $\mathcal{O}(\log |\mathcal{V}|)$ in the first termination case.
Therefore, \ours{} enables fast separations once trained. 
In what follows, we explain how to train \ours{} in detail.

\section{Training \ours{}} \label{sec:training}

Our policy $f_\omega(\mathcal{G})$ predicts the probabilities of whether each vertex is included in the subset $S$ or not.
In other words, we classify the vertices into two sets, $S_t$ and $\mathcal{V}_t \setminus S_t$, at every coarsening step $t$.
We train the policy $f_\omega(\mathcal{G})$ in a supervised manner to learn the exact separation algorithm. \bluenote{Thus, the computational load of the exact separation algorithm is amortized during the training of the policy.}
The labels are collected on relatively small-sized problems, so our model needs to be able to be generalized to larger-sized or out-of-distribution problems.
The exact labels are generated in advance, and the policy is trained to minimize the difference between predictions and the labels.

\subsection{Exact Label Collection} \label{sec:label_gen}
\bluenote{
To obtain separation problems and the exact solutions, we solve CVRP using the cutting plane method with the exact separation algorithm. We generate CVRP instances following \citet{uchoa2017new} and \citet{queiroga202210}. Precisely, the locations of the depot and customers are sampled from the uniform distribution between 0 and 1, and each demand is sampled from a uniform distribution within $[1, 100]$. Because solving separation problems for large-scale CVRP instances with the exact algorithm is computationally demanding, we collect labels by solving relatively small-sized CVRP instances where the number of vertices is within $[50, 100]$.
}

\bluenote{The collection of support graphs and the acquisition of exact labels are conducted offline. Given a relaxed solution $\bar{x}$, the exact separation algorithm solves MIP \crefrange{exact_rci_obj}{exact_rci_c6} for each $M = 0, \ldots, \lceil d(V_C)/Q \rceil - 1$. We use the optimal solution $y^*_i$ as a label of the vertex $i$, i.e., $\hat{\bm{y}} = \{\hat{y}_i\}_{i \in V} = \{ y^*_i\}_{i \in V}$. Even if the set $S^*$, the vertex subset corresponding $y^*$, does not violate the RCI, we include the labels in the dataset. As a result, the policy acquires the ability to identify the vertex set that yields the minimum crossing edge weight. The found RCIs are added to the relaxed problem in order to proceed to the next iteration. This process is reiterated until the algorithm can no longer identify violated RCIs.}

\subsection{Learning with Graph Coarsening} \label{sec:imitation}
We conduct graph embedding and coarsening for batched $N$ support graphs with labels during the training period.
We compute $\hat{q}_{ij}$ via \cref{eq:edge_prob} with labels $\{\hat{y}_i\}_{i \in V}$, it guarantees the labels during the coarsening process.
Note that \bluenote{the model is trained} to imitate labels, so the labels need to be valid when the graphs are coarsened\footnote{For example, two vertices can have a positive contraction probability even though they have different label values. If we merge these two vertices, the label of the merged vertex is ambiguous.}.
This guarantees that the crossing edges of the exact solution are not contracted and, as a result, the minimum crossing edges value is preserved in the coarsening process.
For each coarsening step, a collection of support graph and label pairs $\mathcal{D}_M = \{\mathcal{G}^{(n)}, \hat{\bm{y}}^{(n)}\}_{n=0}^N$ are gathered for each $M \in \mathcal{M}$, where $\mathcal{M} = \{0, \ldots, \max\{K_0, \ldots, K_N\} - 1\}$.

Our policy repeatedly predicts the vertex selection probabilities in the coarsening process and is trained to minimize the difference between predictions and collected labels.
The collected labels are imbalanced depending on $M$, e.g., most vertices have $0$ labels when $M$ is $0$. 
Thus, \bluenote{we employ the Binary Cross-Entropy (BCE) loss function with positive weight as follows:}
\begin{equation}
    \mathcal{L}_M(\omega) = \sum_{(\mathcal{G}, \hat{\bm{y}}) \in \mathcal{D}_M} \sum_{i \in \mathcal{V}} \left[ \rho_M \cdot \hat{y}_i \cdot \log(p_i) + (1 - \hat{y}_i) \cdot \log(1 - p_i) \right],
\end{equation}
where $\rho_M$ is the positive weight of dataset $\mathcal{D}_M$.
\bluenote{The positive weight is computed with the ratio between the number of negative (with a value of 0) and positive (with a value of 1) labels, i.e., }
\begin{equation} \label{eq:pos_weight}
    \rho_M = \frac{\sum_{n=0}^N \sum_{i \in \mathcal{V}_{(n)}} (1 - \hat{y}_{i})}{\sum_{n=0}^N \sum_{i \in \mathcal{V}_{(n)}} \hat{y}_i}.
\end{equation}
It prevents the policy from training towards the dominant value by adjusting the weights as if there are more samples in the less dominant value.
In conclusion, the policy parameters are updated to minimize the training loss defined as follows:
\begin{equation} \label{eq:total_loss}
    \mathcal{L}(\omega) = \sum_{M \in \mathcal{M}} \frac{|\mathcal{D}_M|}{\sum_{m \in \mathcal{M}}|\mathcal{D}_m|} \mathcal{L}_M(\omega).
\end{equation}

\begin{algorithm}
\caption{Training \ours{}}
\label{algo:training}
\small
\begin{algorithmic}[1]
    \Require  Data collection $\mathcal{D} = \{ \mathcal{D}_0, \ldots, \mathcal{D}_{|\mathcal{M}|} \}$, coarsening ratio $\gamma$, max iteration $T$, and learning rate $\alpha$
    \Ensure  Trained parameter $\omega$
    \ForEach {$\{\mathcal{G}^{(n)}, \hat{\bm{y}}^{(n)}\} \in \mathcal{D}$} 
    \State Initialize parameters $\omega$ of the policy
    \State Initialize $G_0 \gets \mathcal{G}^{(n)}$, and $t \gets 0$
    \While{$t \leq T$}
        \State $\{p_i\}_{i \in V_t} \gets f_\omega(G_t)$
        \Comment{Predict vertex selection probability}
        \State Update $\omega \gets \omega + \alpha \argmin_\omega \nabla \mathcal{L}$ \Comment{\cref{eq:total_loss}}
        \State Calculate contraction probabilities $\hat{\bm{q}}$ via $\hat{\bm{y}}^{(n)}$ \Comment{\cref{eq:edge_prob}}
        \If{$|V_{t+1}| \leq 3$ or $\min_{(i, j) \in E_t} \hat{q}_{ij} = 0$}
            \State Terminate the inner loop
        \Else
            \State $G_{t+1} \gets \gamma\textsf{-coarsening}(G_t, \hat{\bm{q}}, \gamma)$ \Comment{\cref{algo:coarsening}}
            \State $t \gets t+1$
        \EndIf
    \EndWhile
    \EndFor
    \\
    \Return $\omega$
\end{algorithmic}
\end{algorithm}

\ours{} is trained with \cref{algo:training}.
As the policy is learned to minimize the weighted BCE loss in \cref{eq:total_loss}, which is the difference between the policy predictions and labels, the crossing edge weights do not need to be calculated.
Thus, the algorithm only conducts the graph coarsening phase without the set assignment and uncoarsening.
To train $f_\omega(\cdot)$, we collect about $20,000$ pairs of the support graph and exact labels from $500$ random CVRP instances.

\section{Experiments} \label{sec:exp}

We evaluate the proposed approach by computing the lower bound of CVRP.
Since CVRPSEP, our baseline, was originally developed for the branch-and-cut algorithm, we also compare CVRPSEP and \ours{} within the cutting plane method to solve the relaxed LP problem at the root node of the branch-and-cut algorithm.\footnote{The source code is available at \url{https://github.com/hyeonahkimm/neuralsep}.}

Experiments are designed to answer the following questions:

\begin{description}%
    \item[] (\textbf{Scalability}) Is the model scalable to larger instances? As the label collections are computationally expensive, we trained our model using relatively small instances with $|V| \in [50, 100]$.
    Therefore, we need to verify whether \ours{} performs well even if the size of problems is larger than the training instances, with $|V| \in (100, 1000]$.
    \item[] (\textbf{Transferability}) Can the model be applied to unseen problems without additional training? 
    The training instances are generated with CVRP whose demands are uniformly distributed in $[1, 100]$. 
    However, the trained model needs to adapt to problems that have different distributions from the original training data.
    \item[] (\textbf{Effectiveness}) Is the graph coarsening effective in solving the separation problems for RCIs?
    We utilize graph coarsening operations to deal with the combinatorial nature of the RCI separation problems.
    We measure violations of cuts found by \ours{} to verify the effectiveness of the proposed algorithm on the separation problems.
\end{description}

\subsection{Experiment Setting} \label{sec:experiment_setting}

\paragraph{Baseline.} 
For the baseline separation algorithm, we use the CVRPSEP library \citep{lysgaard2004new}, written in C.
\bluenote{The library comprises four RCI heuristics -- a connected component algorithm and three shrinking-based heuristics, conducted sequentially when the algorithm fails.
We set the maximum number of RCIs per iteration} as $\min\{|V|, 100\}$ following \cite{lysgaard2004new}.

\textit{Instance generation.}
As described in \cref{sec:label_gen}, the size of CVRP for training is in $[50, 100]$, which is uniformly sampled.
\bluenote{To evaluate the performance of the policy, test instances are generated with the different numbers of customers $|V| \in \{50, 75, 100, 200, 300, 400, 500, 750, 1000\}$. Each test dataset consists of ten instances sampled from the distributions whose locations and demands are uniformly distributed as the same as the training dataset.}
Note that our instance generator follows the one of instance generation logic with random distributions in \citet{queiroga202210}, which originates from \citet{uchoa2017new}.

\paragraph{Implementation.} 
The main cutting plane method is implemented in Julia v1.16 with the LP problems modeled with JuMP.jl v0.21 \citep{DunningHuchetteLubin2017} and solved by CPLEX v12.10.
The CVRPSEP library is called directly in Julia.
While evaluating the trained models, the primary Julia procedure invokes the Python-implemented code using PyCall.jl\footnote{Available at \url{https://github.com/JuliaPy/PyCall.jl}} to solve the separation problems using the trained model. The online supplement provides details regarding the hyperparameters and network architectures used. 
Irrespective of the separation algorithms employed, we follow the same RCI formulation rule with \citet{lysgaard2003cvrpsep}.
For any given set $S$, the rule selects the form with a relatively small number of nonzero coefficients out of three equivalent forms:
\begin{enumerate}[(i)]
    \item $x(S:S) \leq |S| - k(S)$,
    \item $x(\delta(S)) \geq 2k(S)$,
    \item $x(V_C \setminus S:V_C \setminus S) + \frac{1}{2} x(\{0\}:V_C \setminus S) - \frac{1}{2} x (\{0\}:S) \leq |V_C \setminus S| - k(S)$,
\end{enumerate}
where $x(A:B)$ denotes the sum of edge weights whose one end is in $A$ and the other is in $B$.
Form (i) is employed when $|S| \leq {|V|}/{2}$, and form (iii) is employed otherwise.

\paragraph{Performance metric.} Since the cost scale varies depending on the problem instances, we compute the optimality gap to measure the performance of the algorithms. 
The cost lower bounds of each instance are obtained by the cutting plane methods with different separation algorithms.
To compare the performance of the algorithms, we calculate the optimality gap of the resulting lower bound (LB) as follows:
\begin{equation} \label{eq:lb_gap}
    \text{GAP} = \frac{(\text{UB} - \text{LB})}{\text{UB}} \times 100 (\%),
\end{equation}
where UB is computed via the hybrid genetic search (HGS) algorithm of \citet{vidal2022hybrid} for randomly generated instances.

\paragraph{Evaluation.}
\bluenote{The cutting plane method is terminated under two conditions: either when the separation algorithm is unable to identify any violated RCIs or when the number of iterations attains the limit.
We compare the lower bounds and the optimality gap resulting from different separation algorithms, CVRPSEP and NeuralSEP.
Note that comparing the wall clock time of each algorithm directly is difficult because of exogenous factors, including different languages implementing the algorithms, the interface among them, the dependency on external libraries, and other processors. Therefore, we evaluate each algorithm mainly based on the number of iterations.}

\paragraph{Network architecture.}
We use a GNN with five layers for graph embedding, whose vertex and edge update functions employ MLPs with hidden dimensions of $[64, 32]$ and ReLU activation. The probability prediction module is implemented as a simple MLP with a hidden dimension of $[64, 32]$ and ReLU activation and a sigmoid activation function for the final output.

\paragraph{Training parameters.}
We provide the hyperparameters used when we train \ours{} in \cref{table:hyperparams}. We make a batch with 16 separation problems, which contain $K$ problems each, so the total batch size is $\sum_{n=1}^{16} K^{(n)}$.
We use the same hyperparameters except for the coarsening ratio and the maximum coarsening iteration.
We trained our model using PyTorch 1.12.1 on the server equipped with AMD EPYC 7542 CPU and NVIDIA RTX A6000 GPU.

\begin{table}[!ht]
    \caption{Hyperparameter settings}
    \small
    \centering
\begin{tabular}{ll}
    \toprule
    Hyperparameter & Value \\
    \midrule
    Coarsening ratio $\gamma$ & $0.75$ \\
    Maximum coarsening iterations & $50$ \\
    Batch size & $16$ \\
    Number of epochs & $20$ \\
    Optimizer & Adam \\
    Learning rate & $5e^4$ \\
    Learning rate scheduler & CosineAnnealingWarmRestarts \\
    Scheduler $T_0$ & $32$ \\
    \bottomrule
\end{tabular}
    \label{table:hyperparams}
\end{table}

\subsection{Performances on Randomly Generated CVRP} \label{sec:exp_with_random}

\begin{table}
\caption{\bluenote{The comparative results of CVRPSEP and ours within the limited iterations.}}
\label{table:random_instances_rst}
\small
\resizebox{\textwidth}{!}{
\centering
\begin{tabular}{m{1in}|r|rrrrr}
	\toprule
	Method 
	&  Size & Avg. Gap ($\downarrow$) & Avg. LB ($\uparrow$) & Std. Dev. LB &           Avg. Iter. & Avg. $\Delta$ LB \\ \midrule
	\multirow{9}{*}{CVRPSEP}
	&    50 & \textbf{1.968\%} & \textbf{9,363.600} &    2,490.689 & 26 ($\leq$ 200) & \textbf{97.460} \\
	&    75 & \textbf{2.770\%} & \textbf{13,355.075} &    7,286.675 & 40 ($\leq$ 200) & \textbf{104.984} \\
	&   100 & \textbf{4.541\%} & \textbf{15,936.576} &    5,154.639 & 50 ($\leq$ 200) & \textbf{111.857} \\
	&   200 & \textbf{6.281\%}& \textbf{21,378.035} &    4,763.083 & 116 ($\leq$ 200) & \textbf{68.795} \\
	&   300 &  \textbf{8.984\%} & \textbf{31,245.767} &   11,589.540 & 178 ($\leq$ 200) & 70.764 \\
	&   400 & 17.779\% & 40,845.554 &   11,695.275 &          100 ($\leq$ 100) & 173.550 \\
	&   500 & 20.504\% & 47,381.467 &   21,689.405 &          100 ($\leq$ 100) & 187.180 \\
	&   750 & 33.779\% & 61,092.462 &   20,024.117 & 50 ($\leq$ \phantom{1}50) & 475.452 \\
	& 1,000 & 37.195\% & 59,480.493 &   12,122.179 & 50 ($\leq$ \phantom{1}50) & 429.882 \\ \midrule
	\multirow{9}{*}{\shortstack[l]{\ours{} \\(ours)}} 
	&    50 & 4.055\% & 9,129.823 &    2,474.901 &           38 ($\leq$ 200) & 61.473 \\
	&    75 & 5.143\% & 13,064.861 &    7,281.095 &           61 ($\leq$ 200) & 61.263 \\
	&   100 & 6.709\% & 15,593.334 &    5,170.822 &           94 ($\leq$ 200) & 56.308 \\
	&   200 & 9.148\% & 20,742.544 &    4,728.970 &          127 ($\leq$ 200) & 58.559 \\
	&   300 & 10.414\% & {31,092.012} &   12,510.644 &          158 ($\leq$ 200) & \textbf{75.812} \\
	&   400 & \textbf{12.879\%} & \textbf{43,896.118} &   14,509.259 & 100 ($\leq$ 100) & \textbf{204.055} \\
	&   500 & \textbf{14.185\%} & \textbf{53,865.885} &   27,457.718 & 100 ($\leq$ 100) & \textbf{234.865} \\
	&   750 & \textbf{20.243\%} & \textbf{73,652.108} &   23,963.772 & 50 ($\leq$ \phantom{1}50) & \textbf{726.645} \\
	& 1,000 & \textbf{22.857\%} & \textbf{73,140.790} &   15,059.967 & 50 ($\leq$ \phantom{1}50) & \textbf{703.088}  \\ \bottomrule
\end{tabular}
}
\end{table}

\bluenote{This section provides experimental results on the randomly generated CVRP instances. It is noteworthy that the evaluations include a larger size of problems than the training dataset, where the problem size is from 50 to 100.
The lower bound and optimality gap are evaluated based on a restricted number of iterations.
We also compute the average improvement of the lower bound per iteration (Avg. $\Delta$ LB) to compare the quality of the cuts.
The total LB improvements are calculated by the differences between the final LB and the first relaxed cost, which is the minimized cost without considering any capacity constraints.
Lastly, we measure runtime per iteration (Iter. Time) of each algorithm, considering that the iteration limits are varying on the problem size.}
\cref{table:random_instances_rst} shows that \ours{} performs well in larger instances and achieves higher LB than CVRPSEP for $N \geq 400$, despite the training range being $[50, 100]$. The detailed outcomes for individual instances can be found in the supplementary material, specifically in EC.4.

\begin{figure}
\centering
\hspace*{\fill}
\begin{subfigure}{0.465\textwidth}
    \includegraphics[width=\textwidth]{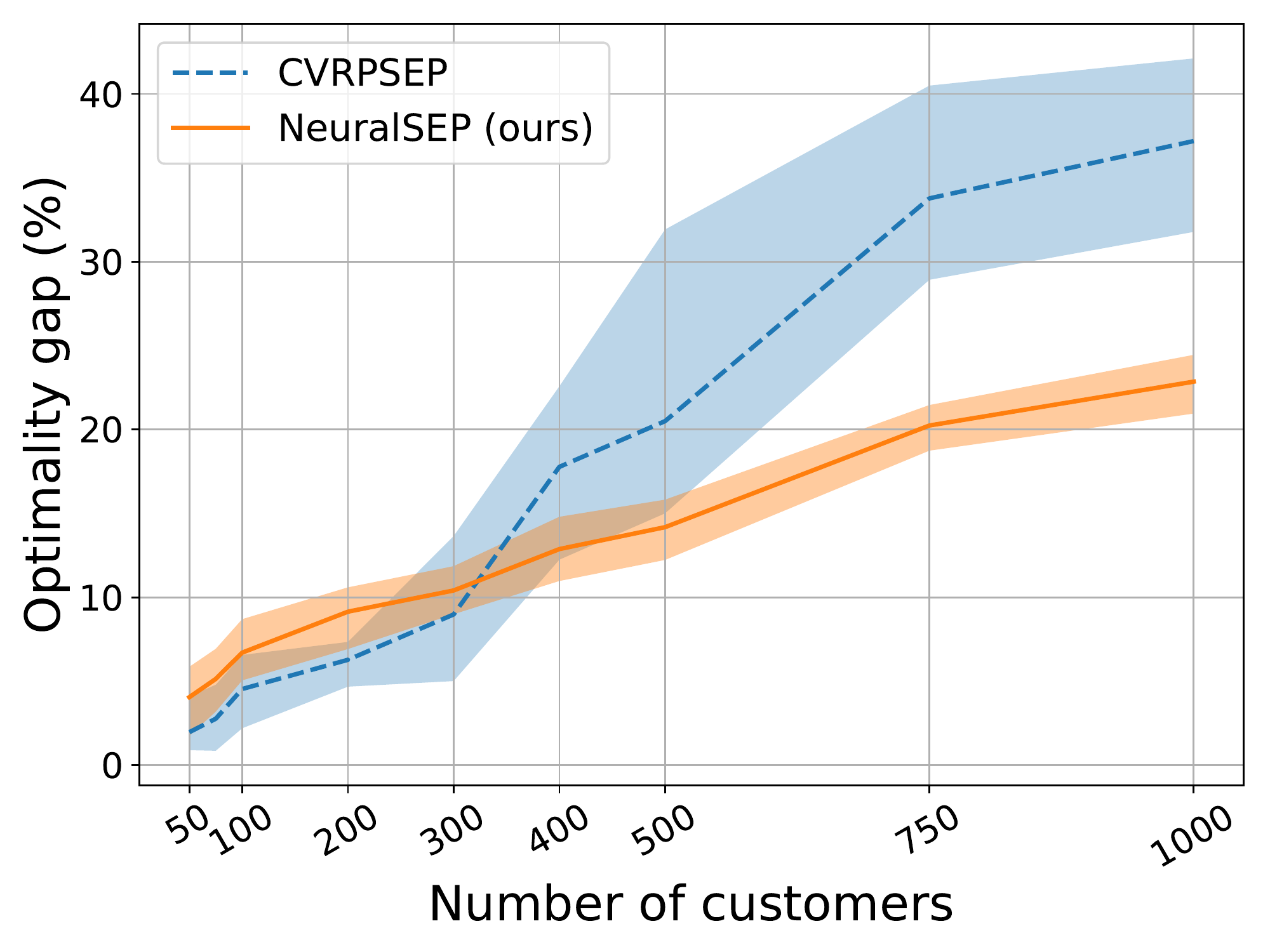}
    \caption{The average and range of the optimality gap.}
    \label{fig:lb_gap_results}
\end{subfigure}
\hspace*{\fill}
\begin{subfigure}{0.465\textwidth}
    \includegraphics[width=\textwidth]{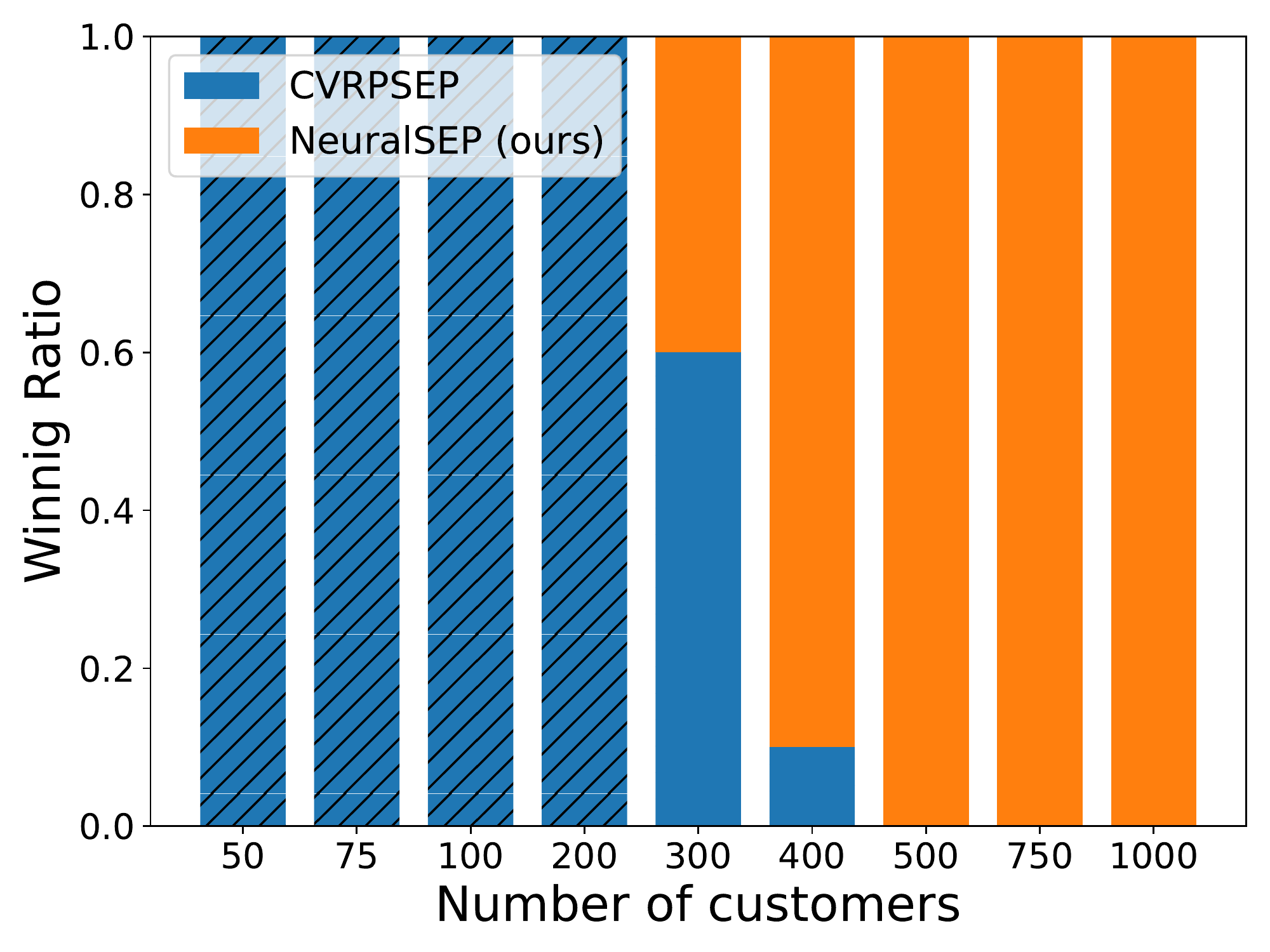}
    \caption{Winning ratio out of 10 instances.}
    \label{fig:winnig_ratio}
\end{subfigure}
\caption{{The results of the cutting plane method with CVRPSEP and \ours{}.}}
\label{fig:random_instances_rst}
\end{figure}

\begin{figure}
  \centering
  \includegraphics[width=0.95\textwidth]{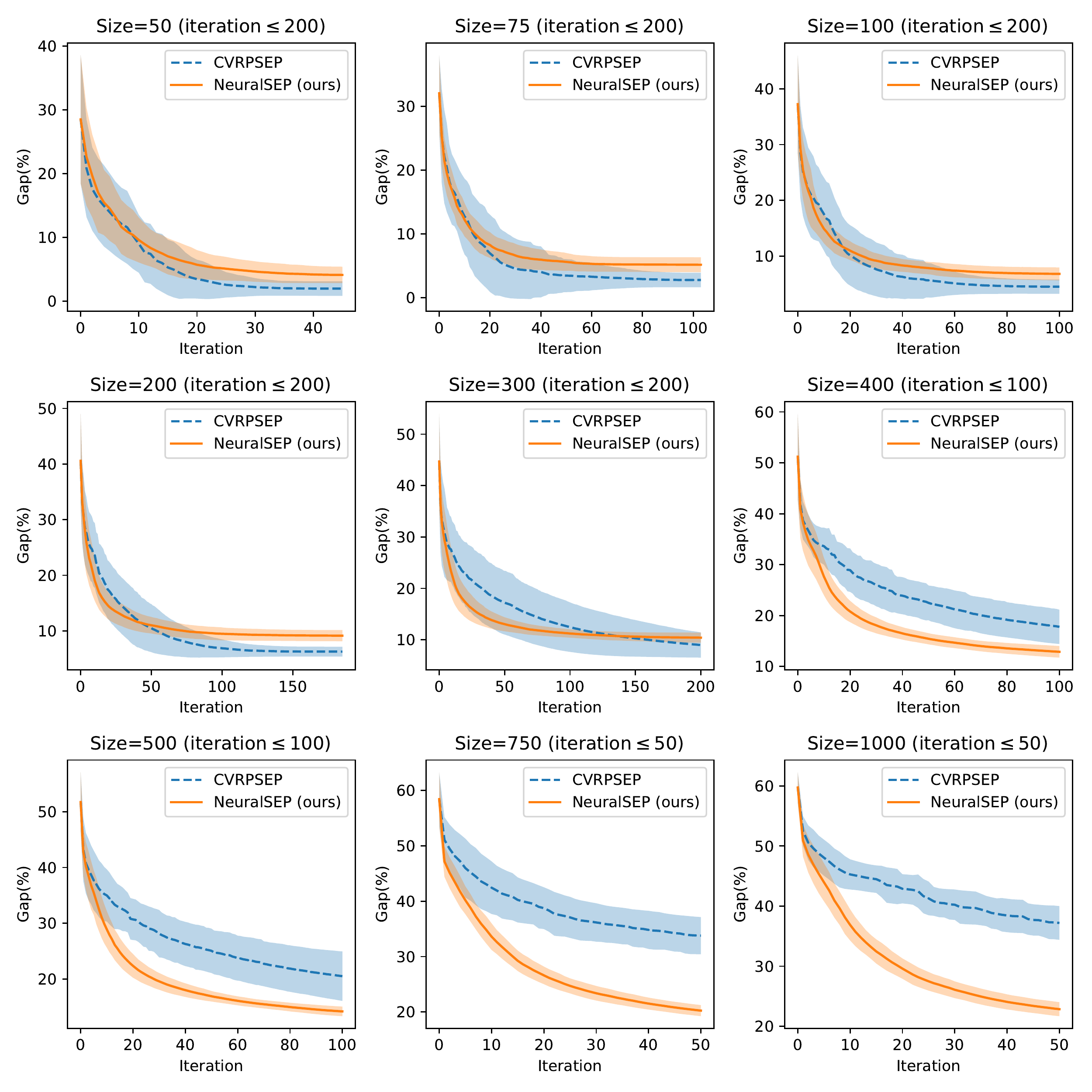}
  \caption{\bluenote{The optimality gap improvement according to iterations. The solid line in the plot denotes the average optimality gap of 10 instances, while the shaded area illustrates the range of the optimality gap.}}
  \label{fig:iteration}
\end{figure}

As the cost scale tends to depend on the size of the problems, we calculate the optimality gap using HGS solutions.
The optimality gap values of CVRPSEP and \ours{} are plotted in \cref{fig:lb_gap_results}.
In addition, \bluenote{the winning ratio (i.e., the number of times defeating each other)} out of $10$ instances is plotted in \cref{fig:winnig_ratio}.
\bluenote{The figures illustrate that our algorithm outperforms CVRPSEP in large problems where} the number of customers is 400 or more.
Furthermore, for the case of $N \geq 500$, \ours{} outperforms CVRPSEP for every instance. 

Despite the noticeable performance, ours tends to exhibit higher time consumption per iteration.
It is noteworthy that CVRPSEP is implemented in C, a programming language renowned for its speed advantages over Python, and the current implementation of \ours{} depends on external libraries that could be engineered further to find cuts time-efficiently. 
To provide a comprehensive comparison, we also undertake experiments within constrained computational time to validate the efficacy of \ours{}, as detailed in \cref{sec:exp_with_time}.

\cref{fig:iteration} illustrates the optimality gap obtained by CVRPSEP and \ours{} with respect to the iterations.
\bluenote{The optimality gap has a notable reduction in the early stages, with diminishing improvements as more cuts are added. 
\cref{fig:iteration} points out that CVRPSEP eventually converges to a lower gap compared to \ours{}.
This tendency can also be seen in \cref{table:random_instances_rst} where the iterations of \ours{} terminate earlier than CVRPSEP when the size is under 400.}
However, when the number of vertices is large, the optimality gap of \ours{} reaches a lower point as compared to that of CVRPSEP within the limited iterations.

\subsection{Performances on X-instances} \label{sec:exp_with_cvrplib}

\begin{table}
\caption{\bluenote{The comparative results of CVRPSEP and ours within the limited iterations on X-instance.}}
\small
\label{table:x_rst}
\resizebox{\textwidth}{!}{
\centering
\begin{tabular}{m{1in}|l|rrrrr}
	\toprule
	Method 
	&  Range & Avg. Gap ($\downarrow$) & Avg. LB ($\uparrow$) & Std. Dev. LB & Avg. Iter. & Avg. $\Delta$ LB \\ \midrule
	\multirow{9}{*}{CVRPSEP}
	& [100, 200) & \textbf{4.271\%} & \textbf{26,030.758} &    14,342.726 & 105 ($\leq$ 200) & \textbf{86.913} \\
	& [200, 300) & \textbf{7.090\%} & \textbf{37,748.285} &    26,060.201 & 173 ($\leq$ 200) & 84.279 \\
	&   [300, 400) & 13.758\% & 47,291.521 &    33,842.847 & 100 ($\leq$ 100) & 186.732 \\
	&   [400, 500) & 16.095\% & 62,615.572 &    48,144.625 & 100 ($\leq$ 100) & 204.960 \\
	&   [500, 600) &  19.527\% & 70,233.048 &   35,998.370 & 50 ($\leq$ \phantom{1}50) & 466.625 \\
	&   [600, 700) & 24.748\% & 63,587.607 &   26,396.767 &          50 ($\leq$ \phantom{1}50) & 422.510 \\
	&   [700, 800) & 30.893\% & 62,362.005 &   29,201.413 &          50 ($\leq$ \phantom{1}50) & 432.440 \\
	&   [800, 900) & 25.495\% & 83,282.409 &   40,126.377 & 50 ($\leq$ \phantom{1}50) & 471.568 \\
	& [900, 1000] & 30.728\% & 106,990.158 &   88,934.935 & 50 ($\leq$ \phantom{1}50) & 781.529 \\ \midrule
	\multirow{9}{*}{\shortstack[l]{\ours{} \\(ours)}} 
	&  [100, 200) & 6.392\% & 25,514.464 &   14,192.554 &          136 ($\leq$ 200) &           60.119 \\
	&  [200, 300) & 8.515\% & 37,660.482 &   26,874.093 & 168 ($\leq$ 200) &           \textbf{84.577}  \\
	&  [300, 400) & \textbf{11.022\%} & \textbf{50,172.212} &   37,826.764 & 100 ($\leq$ 100) &          \textbf{215.539}  \\
	&  [400, 500) & \textbf{11.768\%} & \textbf{66,723.129} &   51,968.514 &          100 ($\leq$ 100) &          \textbf{246.035}  \\
	&  [500, 600) & \textbf{14.449\%} & \textbf{76,593.828} &   42,385.466 & 50 ($\leq$ \phantom{1}50) &          \textbf{593.841}  \\
	&  [600, 700) & \textbf{16.866\%} & \textbf{70,233.887} &   28,931.966 & 50 ($\leq$ \phantom{1}50) &          \textbf{555.436}  \\
	&  [700, 800) & \textbf{19.039\%} & \textbf{71,564.764} &   28,957.874 & 50 ($\leq$ \phantom{1}50) &          \textbf{616.495}  \\
	&  [800, 900) & \textbf{17.601\%} & \textbf{92,101.115} &   44,707.908 & 50 ($\leq$ \phantom{1}50) &          \textbf{647.942} \\
	& [900, 1000] & \textbf{22.832\%} & \textbf{115,312.115} &   85,135.955 & 50 ($\leq$ \phantom{1}50) &          \textbf{947.968} \\ 
 \bottomrule
\end{tabular}
}
\end{table}

\begin{figure}
\centering
\hspace*{\fill}
\begin{subfigure}{0.465\textwidth}
    \includegraphics[width=\textwidth]{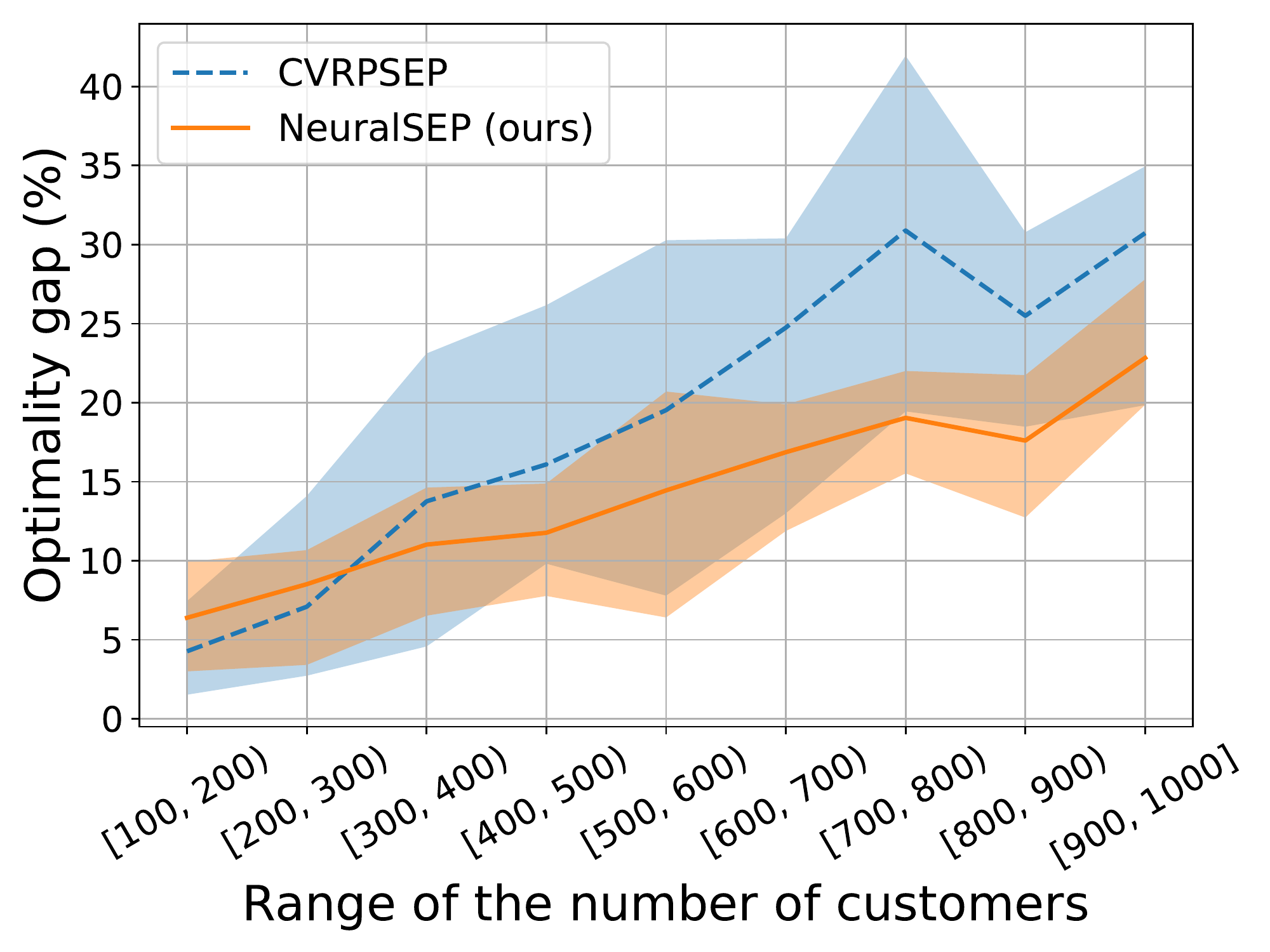}
    \caption{The average and range of the optimality gap.}
    \label{fig:x_gap}
\end{subfigure}
\hspace*{\fill}
\begin{subfigure}{0.465\textwidth}
    \includegraphics[width=\textwidth]{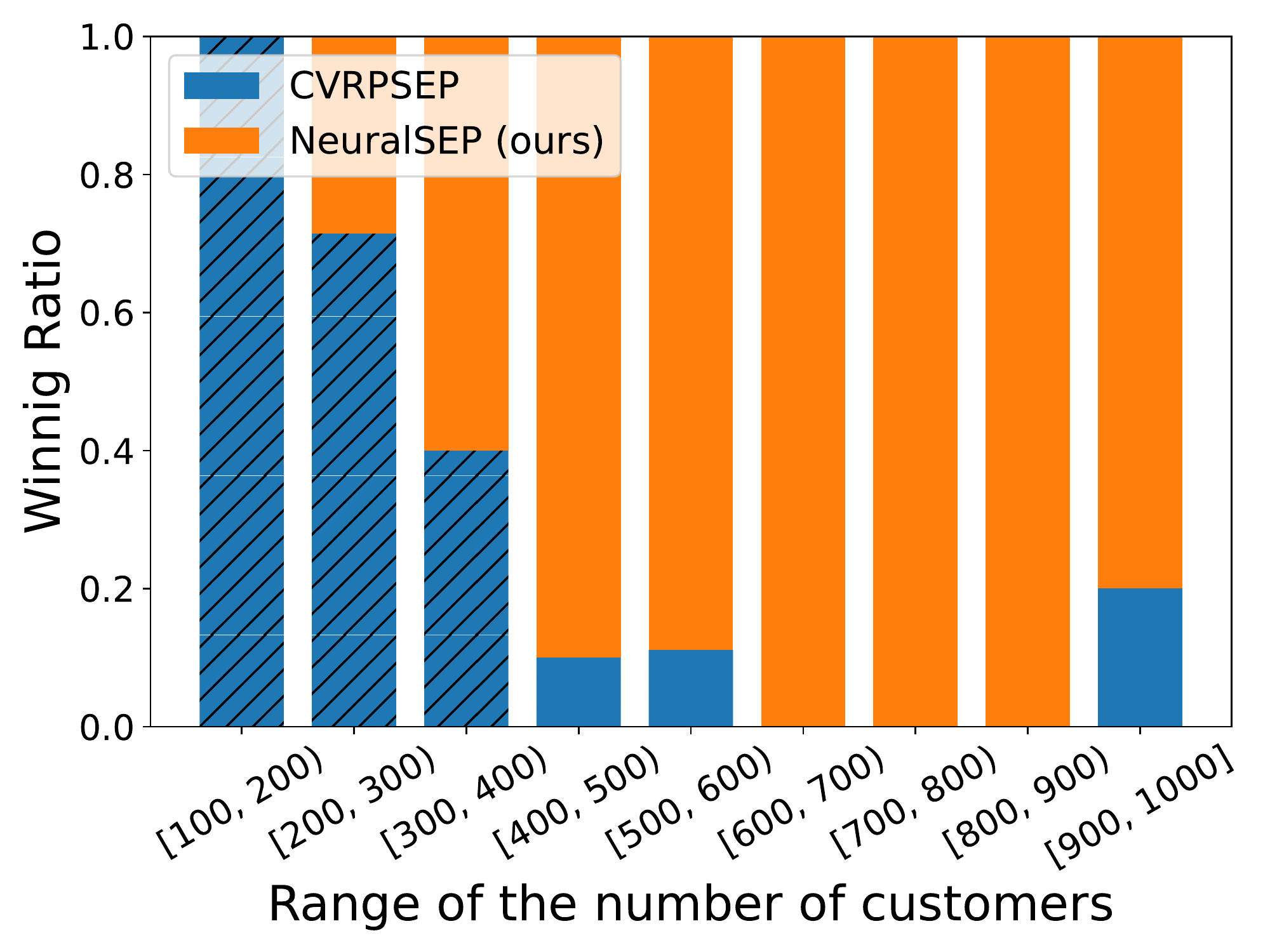}
    \caption{Winning ratio out of 10 instances.}
    \label{fig:x_ratio}
\end{subfigure}
\caption{{The results of the cutting plane method with CVRPSEP and \ours{} in X-instances.}}
\label{fig:x_instances_rst}
\end{figure}

\bluenote{We examine the transferability of our trained model by applying it to X-instances from CVRPLIB \citep{cvrplib} without additional training. \bluenote{Note that demands are sampled from various distributions in X-instances, while our training data employs a uniform demand distribution between 1 and 100.} Our results, as shown in \cref{table:x_rst} and \cref{fig:x_instances_rst}, demonstrate that our model surpasses in large-scale problems with limited iterations, even when the demand distribution differs from the training data. To compute the optimality gap, we use the best-known solutions provided by CVRPLIB.}
The detailed outcomes specific to each instance are provided in EC.4.

In summary, it seems that \ours{} is \emph{implicitly} trained with various demand distributions as the graph coarsening forms different demand distributions by merging the demands in the coarsening phase.
Consequently, \ours{} suffers less from the distribution shifts, which means the performances are less degraded when the test problems are sampled from the different problem distributions with the training problem distribution. 
However, CVRPSEP tends to achieve better performances than our algorithms on X-instances when it is executed enough iterations. In the following section, we provide experimental results to compare the performance of CVRPSEP and \ours{} when the computation time is limited.

\subsection{Results with Limited Computation Time} \label{sec:exp_with_time}

\begin{figure}
\centering
\hspace*{\fill}
\begin{subfigure}{0.45\textwidth}
    \includegraphics[width=\textwidth]{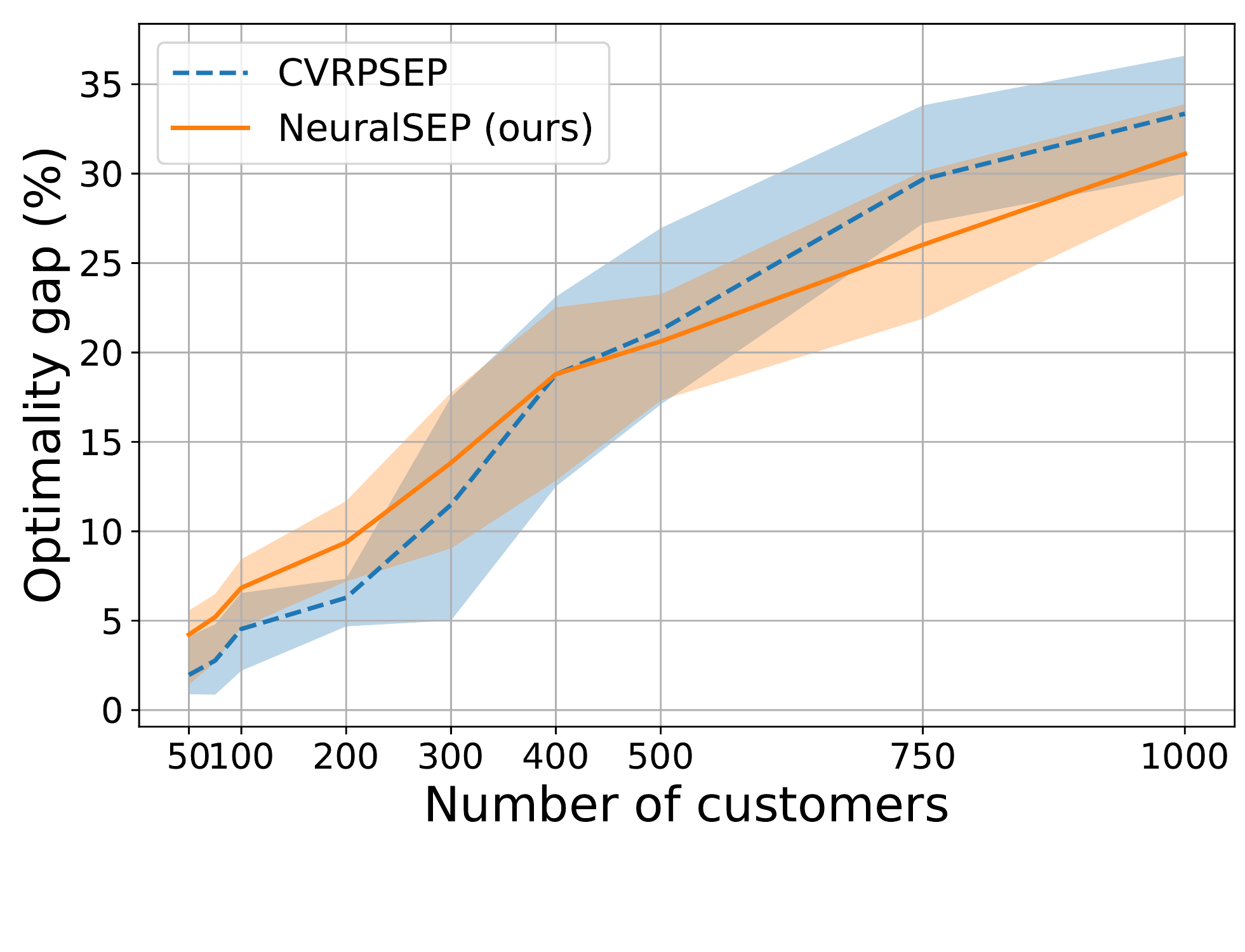}
    \caption{{Randomly generated CVRP (in-distribution)}}
    \label{fig:rand_gap_time}
\end{subfigure}
\hspace*{\fill}
\begin{subfigure}{0.46\textwidth}
    \includegraphics[width=\textwidth]{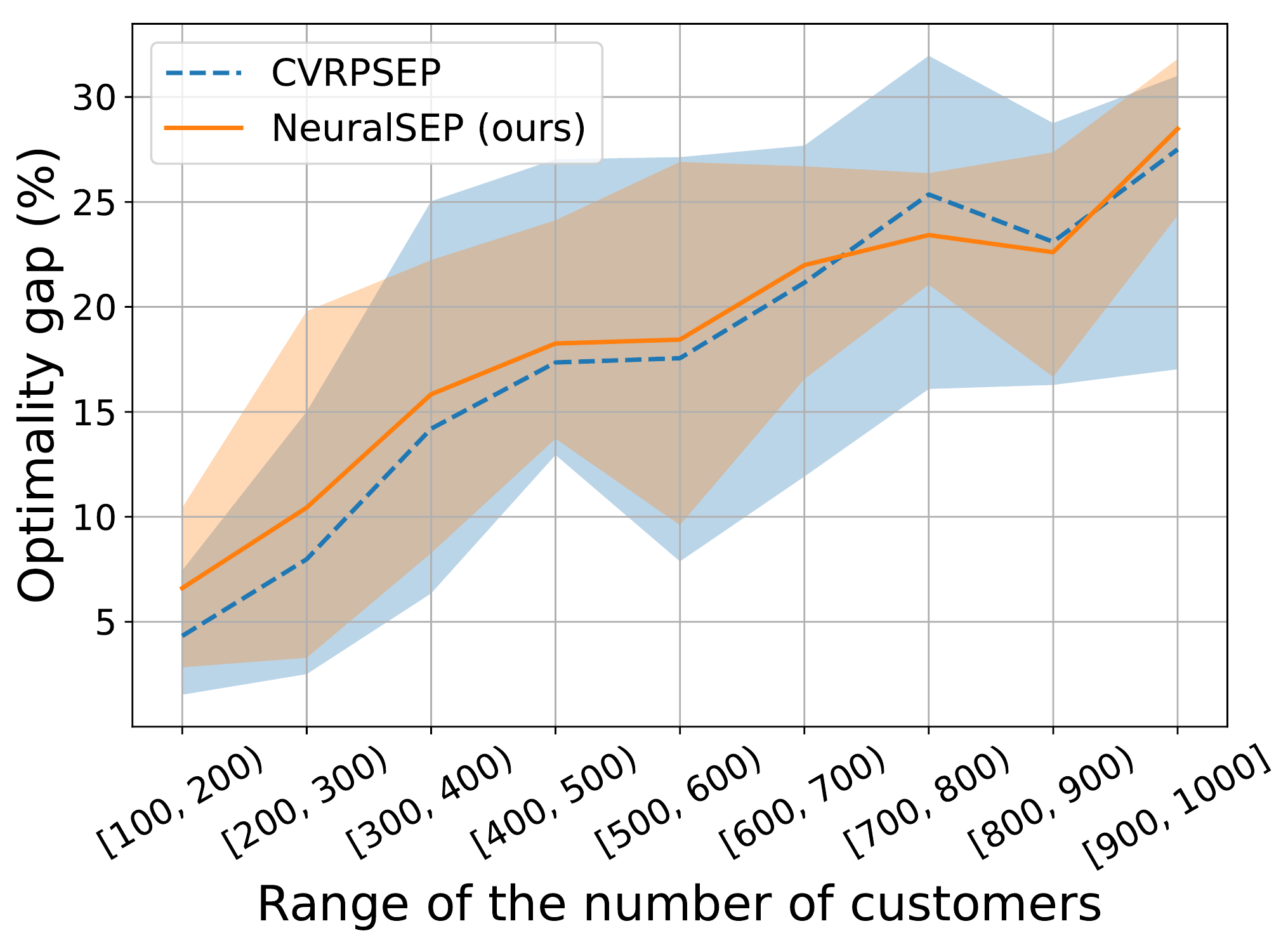}
    \caption{{X-instances (out-of-distribution)}}
    \label{fig:x_gap_time}
\end{subfigure}
\caption{{The range of the optimality gap with 2 hours limit.}}
\label{fig:runtime_exp}
\end{figure}

The better performance within the same number of iterations would not mean practically faster computation.
To test the practical performances, we measure the optimality gap of CVRPSEP and \ours{} with 2 hours limits regardless of the number of iterations, using 4 Intel Xeon Gold 6230 CPUs. 
\ours{} continues to outperform CVRPSEP for problem sizes larger than 400 for randomly generated CVRP instances as depicted in \cref{fig:rand_gap_time}, but exhibits comparable performances in X-instances as shown in \cref{fig:x_gap_time}.
While these trends align with the findings presented in Figure \ref{fig:lb_gap_results} for randomly generated instances, \ours{}'s relatively poorer performance with X-instances can be attributed to its out-of-distribution behavior: our model is trained on the uniform distribution and is directly applied to solve the X-instance which has various distributions (out-of-distribution).
For more comprehensive results, including summary tables and instance-wise time-performance graphs, please refer to the online supplement; see EC.1 and EC.5.

To take advantage of \ours{} fully, we would improve the forward processing time by an efficient implementation and code optimization.
Currently, \ours{} is implemented with the Deep Graph Library \citep[DGL,][]{wang2019deep} in Python.
The widely-used DGL library employs relatively heavy and complex graph data structures to enable various graph operations, while \ours{} utilizes a few simple operations only.
Using a graph data structure specific to \ours{} will expedite the forward processing time.
In addition, we can consider a faster C++ implementation of the graph coarsening procedure, which is currently implemented in slower, native Python. 
The forward processing time can also be significantly improved by using an efficient CPU and GPU integration. 
While GPU will process the inference step certainly faster, we could not justify using GPU due to the expensive overhead between CPU and GPU operations.
Better hardware or code level integration would certainly improve the forward processing time of \ours{}. 

Finding a better lower bound more quickly with \ours{} can be beneficial in two ways.
In most cutting plane methods within the branch-and-cut or branch-price-and-cut algorithms, 
RCIs are applied first, after which other types of cuts---for example, strengthened comb inequalities---are applied.
Since \ours{} converges more quickly, we can employ the other cuts earlier in the cutting plane method, potentially finding tighter bounds more quickly. 
With a better lower bound, we can also reduce the number of branch-and-bound vertices to explore after the cutting plane method. As the branch-and-bound procedure for large-scale problems takes a lot of time, any improvement in the lower bound can be helpful.

\subsection{Performance on Separation Problems} \label{sec:separation_exp}

In this section, we provide experimental results for the separation problems. The test data is collected using the exact separation algorithm in advance. We measure the average violations, the average number of inequalities found by each separation algorithm, and the success rate, which indicates how many times the separation algorithm succeeds in finding at least one RCI.

As shown in \cref{table:violation}, the average of violations observed in NeuralSEP lies between the exact separation algorithm and CVRPSEP. The small average violations observed with CVRPSEP can be attributed to its propensity for generating a multitude of inequalities in comparison to alternative separation algorithms. CVRPSEP gives more than five times of others. Note that the exact separation and NeuralSEP only find inequalities as many as the number of vehicles at maximum. Though ours can find more violated cuts on average than CVRPSEP, our method achieves the lowest success rate. Consequently, NeuralSEP usually converges to the higher optimality gap when NeuralSEP is embedded into the cutting plane method. It can be interpreted as NeuralSEP is less effective if the computation time is enough or the problem sizes are relatively small.
Further analysis for extensive experiments is also provided in the online supplement (EC.2 and EC.3).

\begin{table}
\caption{Separation results of different separation algorithms on various sizes of problems.}
\small
\centering
\label{table:violation}

\begin{tabular}{rrrrrrrrrr} 
\toprule
\multicolumn{1}{c}{Size} & \multicolumn{3}{c}{Avg. Violations}  & \multicolumn{3}{c}{Num. of Inequalities} & \multicolumn{3}{c}{Success Rate} \\ \cmidrule[0.5pt](lr{0.2em}){2-4}\cmidrule[0.5pt](lr{0.2em}){5-7} \cmidrule[0.5pt](lr{0.2em}){8-10}
 & Exact & CVRPSEP & NeuralSEP & Exact & CVRPSEP & NeuralSEP & Exact & CVRPSEP & NeuralSEP \\
\midrule
50 & 1.5640 & 0.2029 & 1.4579 & 547 & 3,158 & 547 & 1.00 & 0.88 & 0.68 \\
75 & 2.5204 & 0.0938 & 2.3529 & 963 & 5,522 & 963 & 0.96 & 0.82 & 0.56 \\
100 & 1.9115 & 0.0759 & 1.8034 & 1,059 & 7,742 & 1,059 & 0.99 & 0.90 & 0.49 \\
200 & 1.1899 & 0.0377 & 0.8764 & 1,263 & 8,934 & 1,263 & 1.00 & 0.89 & 0.34 \\
\bottomrule
\end{tabular}
\end{table}

\section{Conclusion} \label{sec:conclusion}
\bluenote{This study suggests enhancing the cutting plane method for RCIs by employing the neuralized separation algorithm, called \ours{}.}
Even though the cutting plane method is one of the successful methods to solve many CO problems, the performance highly depends on the separation algorithm.
The separation problem is a CO problem known as $\mathcal{NP}$-hard, so the exact separation algorithm requires massive computation. 
Therefore, several human-designed heuristics have been introduced and proved their practicality, but they often find cuts that need to be tighter.

\bluenote{We utilize graph coarsening to discretize the continuous prediction of the neural network model.}
\ours{} iteratively reduces the size of the problem with graph coarsening and re-predicts the assignment probabilities on the coarser graph.
The contracted edges are excluded from the final crossing edge set, so the coarsening process can be interpreted as gradual decision-making for excluding edges.
Our model effectively learns the exact separation algorithm with $\mathcal{O}(|V||E| \log |V|)$ worst-case time complexity for inferencing.

The experiments show that \ours{} finds tighter cuts at each iteration than CVRPSEP, a competitive heuristic library.
Though the model is trained using instances whose customers range from $[50, 100]$ with uniformly distributed demands, it is scalable to larger-sized or new instances with unseen demand distributions.
Moreover, ours outperforms CVRPSEP for problems whose size is more than {$400$} when the number of separation iterations is limited.
These results indicate that \ours{} would be useful for finding exact solutions for large-scale CVRPs with 400 or more.

Incorporating \ours{} within the state-of-the-art branch-price-and-cut algorithms and developing neural separation algorithms for other types of cuts will be important for future research topics to measure the practical implications of \ours{}.

\section*{Acknowledgments}
This research was partially supported by the Institute of Information \& Communications Technology Planning \& Evaluation (IITP) grant funded by the Korean government (MSIT) (Project No. 2022-0-01032) and the National Research Foundation of Korea (Project No. 1711199394).

\bibliographystyle{apalike}
\bibliography{library}

\clearpage
\appendix
\appendixpage
\renewcommand{\appendixpagename}{Appendix}

\label{sec:proofs}

\proof{Proof of \cref{prop:error}}
    Suppose the coarser graph as $\tilde{\mathcal{G}} = (\tilde{\mathcal{V}}, \tilde{\mathcal{E}})$ is obtained by contracting edge $(u, v)$ from $\mathcal{G} = (\mathcal{V}, \mathcal{E})$.
    We need to prove $\sum_{(i, j) \in \delta(S)} \bar{x}_{ij} = \sum_{(i, j) \in \delta (\tilde{S})} \bar{x}_{ij}$ and $\sum_{i \in S} d_i = \sum_{i \in \tilde{S}}d_i$, where $S$ and $\tilde{S}$ are the sets consisting of the vertices with $\hat{y}_i = 1$.
    
    First, we show that contracting edge $(u, v)$ preserves the crossing edge weights and total demand, if $\hat{y}_u = \hat{y}_v$ (i.e., the edge $(u, v)$ is none of the crossing edges).
    We denote $\tilde{v}$ as a super-vertex, i.e., $u$ and $v$ are merged into $\tilde{v}$, with $\hat{y}_{\tilde{v}} = \hat{y}_u = \hat{y}_v$.
    The demand of $\tilde{v}$ is set to $d_u + d_v$ and the demand of other vertices remain the same, so $\sum_{i \in \mathcal{V}} d_i \hat{y}_i = \sum_{i \in \tilde{\mathcal{V}}}d_i \hat{y}_i$; thus, $\sum_{i \in S} d_i = \sum_{i \in \tilde{S}}d_i$.
    The edges connected to $\tilde{v}$ are updated as follows:
    \begin{equation} \label{eq:edge_weight}
        \bar{x}_{\tilde{v}l} = \begin{cases}
        \bar{x}_{ul} + \bar{x}_{vl} &\quad \mbox{if } l \in \mathcal{N}(u) \cap \mathcal{N}(v) \\
        \bar{x}_{ul} &\quad \mbox{if } l \in \mathcal{N}(u) \setminus \mathcal{N}(v) \\
        \bar{x}_{vl} &\quad \mbox{if } l \in \mathcal{N}(v) \setminus \mathcal{N}(u)
        \end{cases}
    \end{equation}
    Since the $(u, v) \notin \delta(S)$, removing $(u, v)$ does not affect the crossing edge weights.
    If $(u, l) \in \delta(S)$ or $(v, l) \in \delta(S)$ (i.e., $\hat{y}_l \neq \hat{y}_u = \hat{y}_v $), then $(\tilde{v}, l) \in \delta(\tilde{S})$ and the weight of the $(\tilde{v}, l)$ is preserved by \cref{eq:edge_weight}. The edges disconnected with $\tilde{v}$ remain the same; thus, $\sum_{(i, j) \in \delta(S)} \bar{x}_{ij} = \sum_{(i, j) \in \delta (\tilde{S})} \bar{x}_{ij}$.
    
    Second, we show that the crossing edges are not selected to contract if $\max_i \epsilon_i < 1/2$.
    Let $\epsilon_i$ be the prediction error for vertex $i$. Then, we can compute $q_{ij}$ with $p_i = |\hat{y}_i - \epsilon_i|$ as follows:
    \begin{align*}
        q_{ij} = \begin{cases}
        (1 - \epsilon_i)(1 - \epsilon_j) + \epsilon_i \epsilon_j, & \quad \mbox{if } \hat{y}_i = 1, \hat{y}_j = 1 \\
        \epsilon_i (1 - \epsilon_j) + (1 - \epsilon_i) \epsilon_j, & \quad \mbox{if } \hat{y}_i = 0, \hat{y}_j = 1 \\
        (1 - \epsilon_i) \epsilon_j + \epsilon_i (1 - \epsilon_j), & \quad \mbox{if } \hat{y}_i = 1, \hat{y}_j = 0 \\
        \epsilon_i \epsilon_j + (1 - \epsilon_i) (1 - \epsilon_j), & \quad \mbox{if } \hat{y}_i = 0, \hat{y}_j = 0 
        \end{cases}
    \end{align*}
    Since we select the edge to contract greedily according to $q_{ij}$, the end points of the selected edge have the same value of $\hat{y}$, when
    \begin{equation} \label{eq:preserving}
    \begin{split}
        &\quad 1 - \epsilon_i - \epsilon_j +2\epsilon_i \epsilon_j > \epsilon_i + \epsilon_j - 2\epsilon_i \epsilon_j \\
        \iff  &\quad 4\epsilon_i \epsilon_j - 2\epsilon_i  -2 \epsilon_j + 1 > 0 \\
        \iff  &\quad \Big(\epsilon_i - \frac{1}{2}\Big) \Big(\epsilon_j - \frac{1}{2}\Big) > 0.
    \end{split}
    \end{equation}
    If the vertex prediction error $\epsilon_i$ is bounded to $1/2$ (i.e., $\max_{i\in \mathcal{V}_t} \epsilon_i < 1/2$), the condition \cref{eq:preserving} is satisfied.
    As the crossing edges are not contracted, the crossing edge weights and the total demand are preserved.
\hfill$\square$
\endproof

\proof{Proof of \cref{prop:complexity}}
    We analyze the time complexity for each part of the forward propagation described in \cref{sec:method}. 
    \begin{itemize}
        \item \textit{Graph embedding:} We employ message passing GNN whose time complexity is known as $\mathcal{O(|E|)}$ \citep{wu2020comprehensive}.
        \item \textit{Graph coarsening:} Computing the $t$-th coarse graph requires $(1 - \gamma) \gamma^{t-1} |\mathcal{V}|$ times of edge contractions (see \cref{algo:coarsening}). A single edge contraction consists of argmax operations for the edges, and sum operations for the selected vertices and their connected edges; thus, it is bounded to $\mathcal{O(|E|)}$. The time complexity of coarsening is bounded to $\mathcal{O}(\gamma^{t-1} (1 - \gamma) \mathcal{|V| |E|)} = \mathcal{O(|V||E|)}$.
        \item \textit{Set assignment:} Set assignment conducts simple rounding and argmax operations for scalar vertex values, so it has $\mathcal{O(|V|)}$ worst time complexity.
        \item \textit{Graph uncoarsening:} As we keep tracking the vertex information in the coarsening phase, we can directly map the coarsened vertices to the original vertices. Therefore, the time complexity of uncoarsening is bounded to $\mathcal{O(|V|)}$.
    \end{itemize}

    Coarsening conducts $T$ iterations of feature embedding and graph coarsening.
    After $T$ iterations, we get the coarsest graph with three vertices (i.e., $|\mathcal{V}_T| = \lfloor \gamma^T |\mathcal{V}| \rfloor = 3$), so the number of iterations is bounded to $\mathcal{O}(\log |\mathcal{V}|)$. 
    \bluenote{The coarsening process terminates when 1) the number of vertices is three or 2) every edge has zero contraction probability. Since the graph is coarsened with ratio $\gamma$, the number of vertices at iteration $t$ is $|\mathcal{V}_t|=\lfloor \gamma |\mathcal{V}_{t-1}| \rfloor \leq \gamma^{t}|\mathcal{V}|$. Thus, the number of iterations is bounded by $\mathcal{O}(\log |\mathcal{V}|)$. 
The second case is an early stopping criterion.
For example, at iteration $t$, if there is only one possible edge to be contracted, but $(1 - \gamma) \gamma^{t-1} |\mathcal{V}| > 1$, then it means the other edge values are zero. Therefore, the coarsening process terminates early, which means the number of iterations is less than in the first case. Accordingly, the number of iterations is bounded by $\mathcal{O}(\log |\mathcal{V}|)$ in both cases. }
Thus, the total time complexity is
    \begin{align*}
        \sum_{t=0}^T \mathcal{O}(|\mathcal{E}_t| + |\mathcal{V}_t||\mathcal{E}_t|) + \mathcal{O}(|\mathcal{V}_t|) + \mathcal{O}(|\mathcal{V}_t|) = \sum_{t=0}^T \mathcal{O}(|\mathcal{V}_t||\mathcal{E}_t|) =
        \mathcal{O}\left(T \left( \mathcal{|V||E|} \right) \right) 
        =  \mathcal{O}(\mathcal{|V||E|} \log |\mathcal{V}|),
    \end{align*}
which completes the proof.
\hfill$\square$
\endproof

 \clearpage
 
\section{Computation Time} \label{sec:runtime_rst}

\subsection{Detailed Results with Limited Computation Time}
The detailed results with the limited computation time of 2 hours are provided in \cref{table:random_runtime,table:x_runtime}.

\begin{table}[!ht]
\caption{The average lower bound of randomly generated instances in 2 hours.}
\small
\label{table:random_runtime}

\centering
\begin{tabular}{m{0.9in}|r|rrrrr}
	\toprule
	Method 
	&  Size & Avg. Gap ($\downarrow$) & Avg. LB ($\uparrow$) & Avg. Runtime &           Avg. Iter. & Winning Ratio \\ \midrule
	\multirow{9}{*}{CVRPSEP}
	&    50 & \textbf{1.968\%} & \textbf{9,363.659} &    0.42 & 26  & 1.0 \\
	&    75 & \textbf{2.770\%} & \textbf{13,355.294} &    5.53 & 40 & 1.0 \\
	&   100 & \textbf{4.535\%} & \textbf{15,937.833} &    5.53 & 52 & 1.0 \\
	&   200 & \textbf{6.283\%}& \textbf{21,377.624} &    1,260.95 & 118 & 1.0 \\
	&   300 &  \textbf{11.486\%} & \textbf{30,140.539} &   6,854.07 & 120 & 0.8 \\
	&   400 & 18.767\% & 40,421.297 &   7,361.86 &          96 & 0.5 \\
	&   500 & 21.255\% & 47,447.202 &   7,468.32 &          103 & 0.4 \\
	&   750 & 29.682\% & 64,821.892 &   7,376.40 & 133 & 0.1 \\
	& 1,000 & 33.353\% & 63,031.672 &   7,575.62 & 101 & 0.2 \\ \midrule
	\multirow{9}{*}{\shortstack[l]{\ours{} \\(ours)}} 
	&    50 & 4.222\% & 9,146.218 &  25.59 & 41 & 0.0 \\
	&    75 & 5.204\% & 13,063.576 & 69.54 & 62 & 0.0 \\
	&   100 & 6.840\% & 15,577.428 & 153.06 & 96 & 0.0 \\
	&   200 & 9.382\% & 20,688.686 & 1,933.56 & 123 & 0.0 \\
	&   300 & 13.832\% & 29,549.060 & 5,888.07 & 69 & 0.2 \\
	&   400 & \textbf{18.785\%} & \textbf{40,558.797} & 7,395.57 & 30 & 0.5 \\
	&   500 & \textbf{20.617\%} & \textbf{47,784.854} & 7,502.68 & 26 & 0.4 \\
	&   750 & \textbf{26.022\%} & \textbf{67,807.911} & 7,550.21 & 22 & 0.9 \\
	& 1,000 & \textbf{31.107\%} & \textbf{65,140.072} & 7,544.50 & 17 & 0.8  \\ \bottomrule
\end{tabular}

\end{table}

\begin{table}[!ht]
\caption{The average lower bound of X-instances in 2 hours.}
\small
\label{table:x_runtime}

\centering
\begin{tabular}{m{0.9in}|r|rrrrr}
	\toprule
	Method 
	&  Range & Avg. Gap ($\downarrow$) & Avg. LB ($\uparrow$) & Avg. Runtime &           Avg. Iter. & Winning Ratio \\ \midrule
\multirow{9}{*}{CVRPSEP}
 & [100, 200) &  \textbf{4.324\%}  &  \textbf{24,565.945}  &  719.96  & 102 & 1.00 \\ 
 & [200, 300) &  \textbf{7.979\%}  &  \textbf{38,046.310}  &  5,727.53  & 151 & 0.77 \\ 
 & [300, 400) &  \textbf{14.196\%}  &  \textbf{47,360.608}  &  7,349.83  & 89 & 0.67 \\ 
 & [400, 500) &  \textbf{17.354\%}  &  \textbf{62,581.232}  &  7,421.87  & 101 & 0.70 \\ 
 & [500, 600) &  \textbf{17.556\%}  &  \textbf{72,695.153}  &  7,480.25  & 106 & 0.78 \\ 
 & [600, 700) &  \textbf{21.157\%}  &  \textbf{66,344.396}  &  7,408.76  & 118 & 0.67 \\ 
 & [700, 800) &  25.358\%  &  66,440.507  &  7,407.79  & 116 & 0.17 \\ 
 & [800, 900) &  23.101\%  &  \textbf{86,058.138}  &  7,413.13  & 106 & 0.33 \\ 
 & [900, 1000] &  \textbf{27.505\%}  &  \textbf{111,911.010}  &  7,391.01  & 90 & 0.60 \\ 
\midrule
\multirow{9}{*}{\shortstack[l]{\ours{} \\(ours)}} 
 & [100, 200) &  6.602\%  &  24,024.549  &  1,463.85  & 140 & 0.00 \\ 
 & [200, 300) &  10.439\%  &  37,041.290  &  6,238.87  & 107 & 0.23 \\ 
 & [300, 400) &  15.835\%  &  46,065.654  &  7,245.67  & 51 & 0.33 \\ 
 & [400, 500) &  18.257\%  &  61,082.248  &  7,573.36  & 24 & 0.30 \\ 
 & [500, 600) &  18.442\%  &  72,135.238  &  7,456.23  & 23 & 0.22 \\ 
 & [600, 700) &  21.992\%  &  65,606.320  &  7,779.46  & 22 & 0.33 \\ 
 & [700, 800) &  \textbf{23.424\%}  &  \textbf{67,376.144}  &  7,597.79  & 24 & 0.83 \\ 
 & [800, 900) &  \textbf{22.606\%}  &  85,758.536  &  7,637.42  & 23 & 0.67 \\ 
 & [900, 1000] &  28.483\%  &  107,405.257  &  7,643.43  & 18 & 0.40 \\

\bottomrule
\end{tabular}

\end{table}

\subsection{Performance-Time Comparison on Randomly Generated CVRP}
\begin{figure}
\centering
\begin{subfigure}{\textwidth}
    \includegraphics[width=0.9\textwidth]{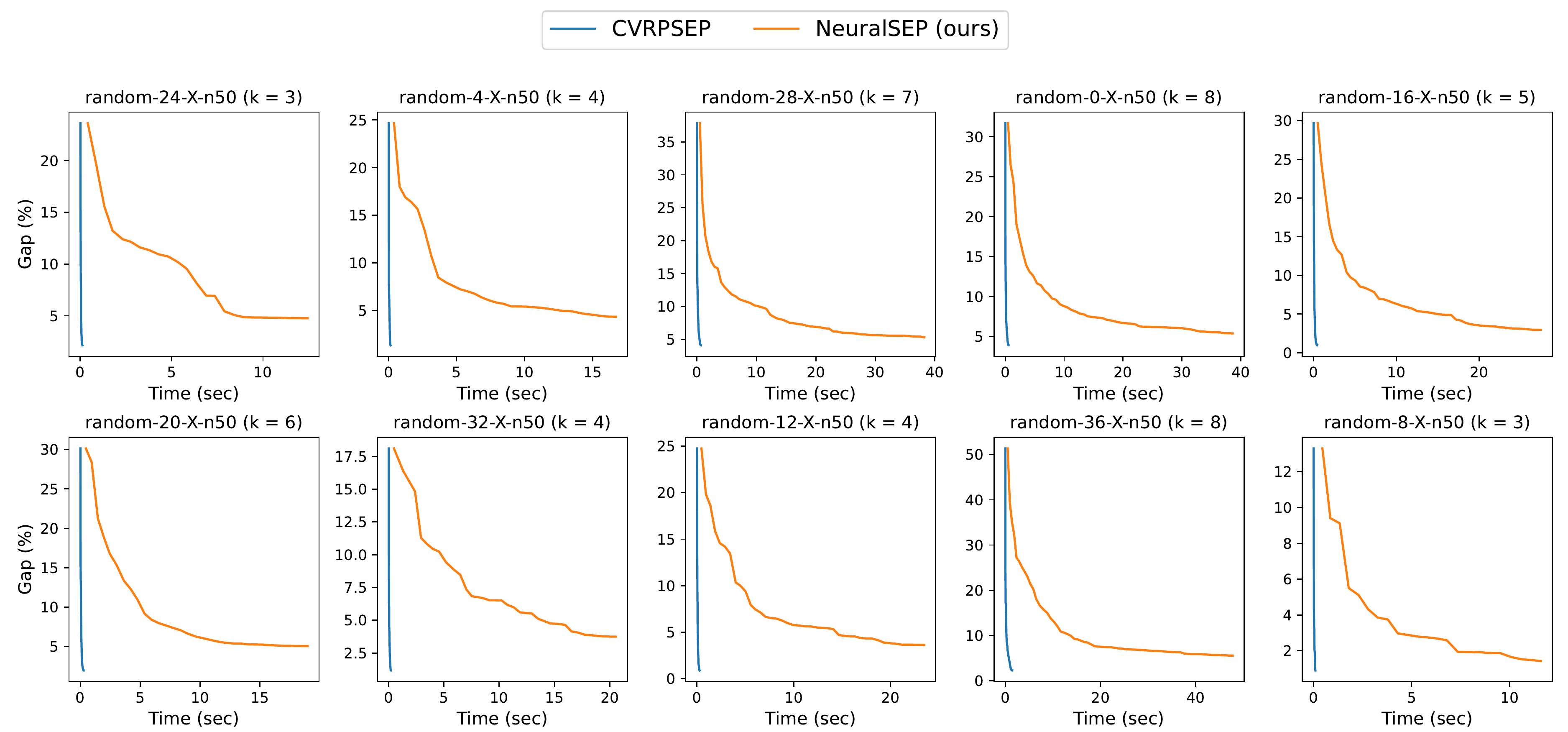}
    \caption{$N = 50$}
    \label{fig:runtime_50}
\end{subfigure}
\hspace*{\fill}
\begin{subfigure}{\textwidth}
    \includegraphics[width=0.9\textwidth]{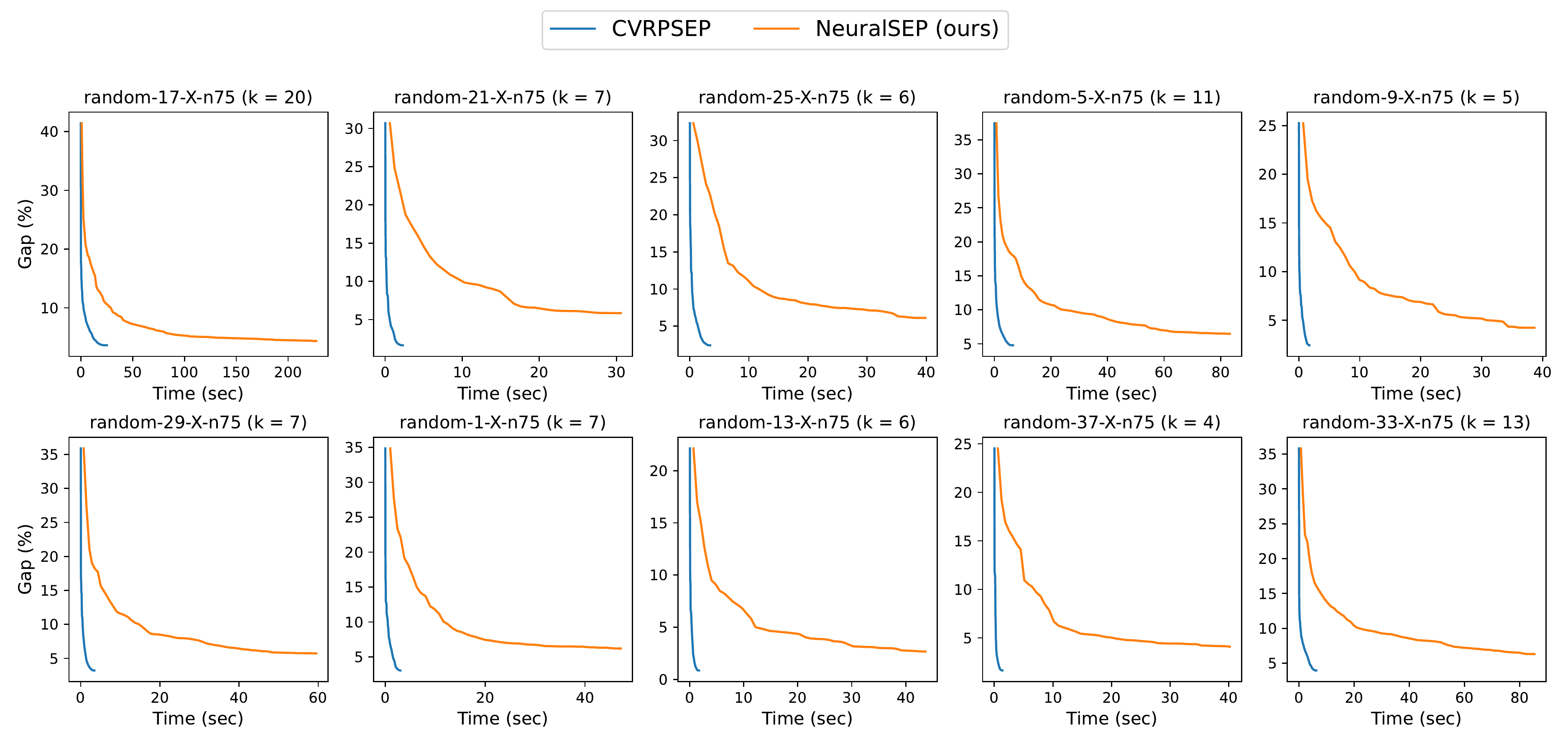}
    \caption{$N = 75$}
    \label{fig:runtime_75}
\end{subfigure}
\hspace*{\fill}
\begin{subfigure}{\textwidth}
    \includegraphics[width=0.9\textwidth]{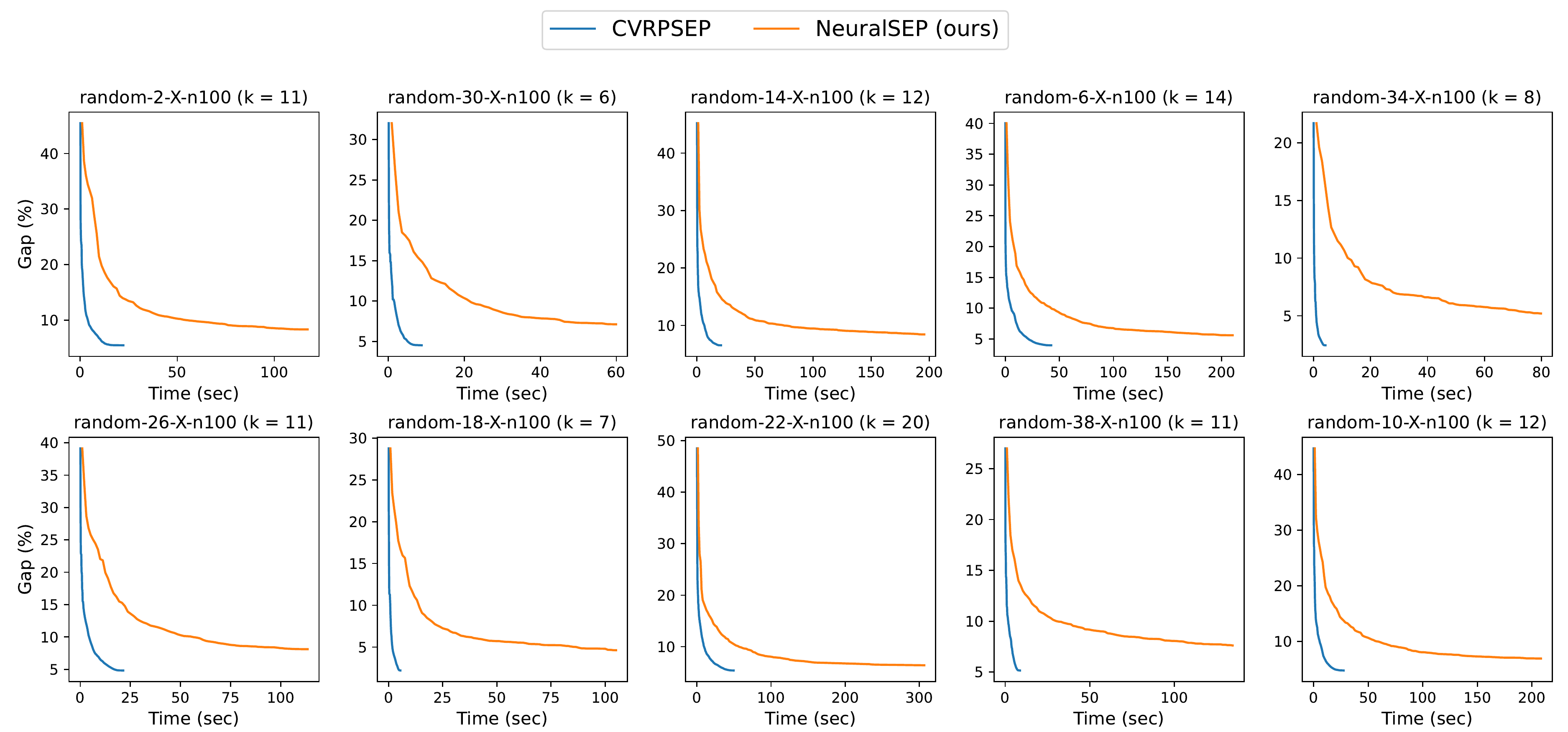}
    \caption{$N = 100$}
    \label{fig:runtime_100}
\end{subfigure}
\caption{Optimality gap comparisons on each instance from $N = 50$ to $N = 100$ with limited time (2h).}
\label{fig:runtime1}
\end{figure}

\begin{figure}
\centering
\begin{subfigure}{\textwidth}
    \includegraphics[width=0.9\textwidth]{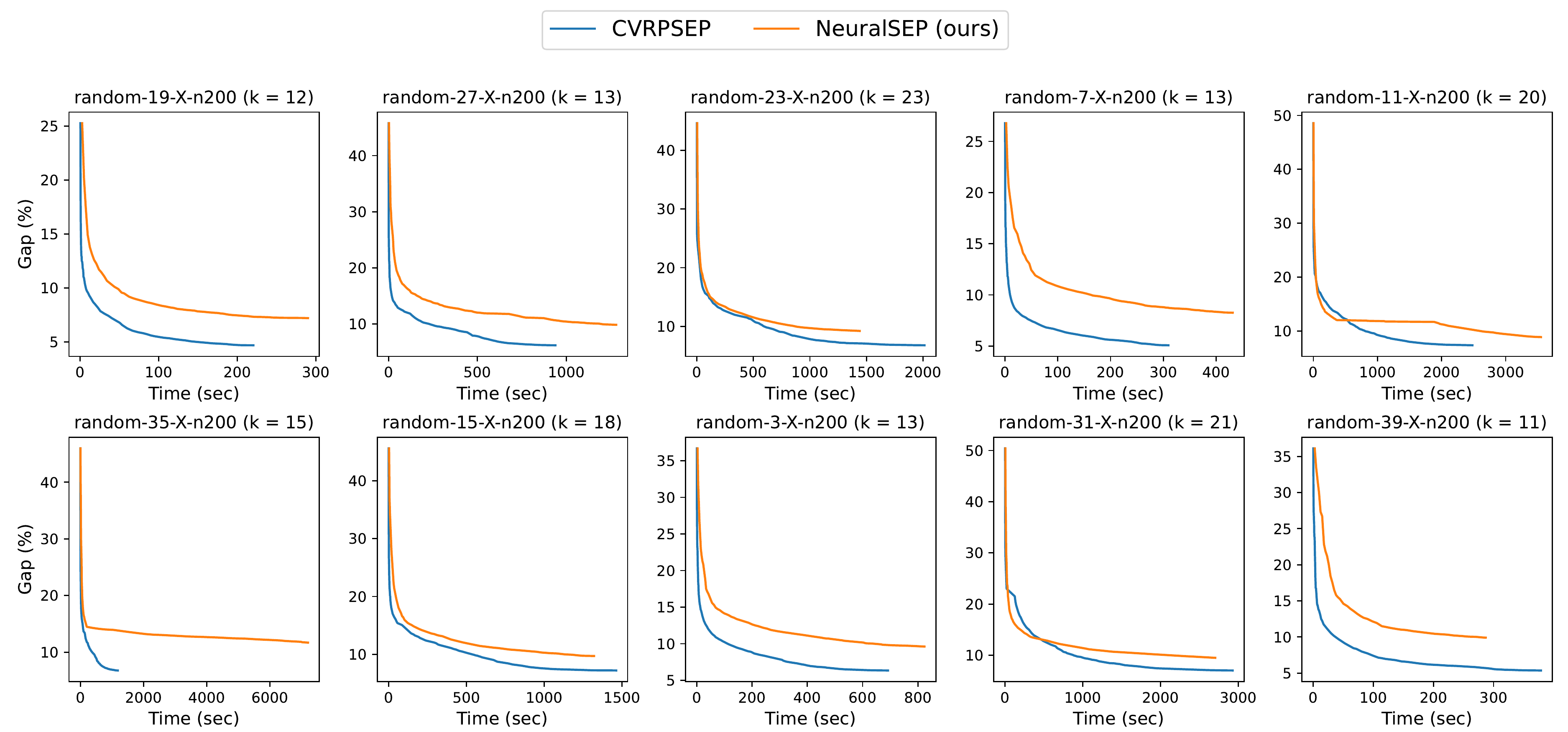}
    \caption{$N = 200$}
    \label{fig:runtime_200}
\end{subfigure}
\hspace*{\fill}
\begin{subfigure}{\textwidth}
    \includegraphics[width=0.9\textwidth]{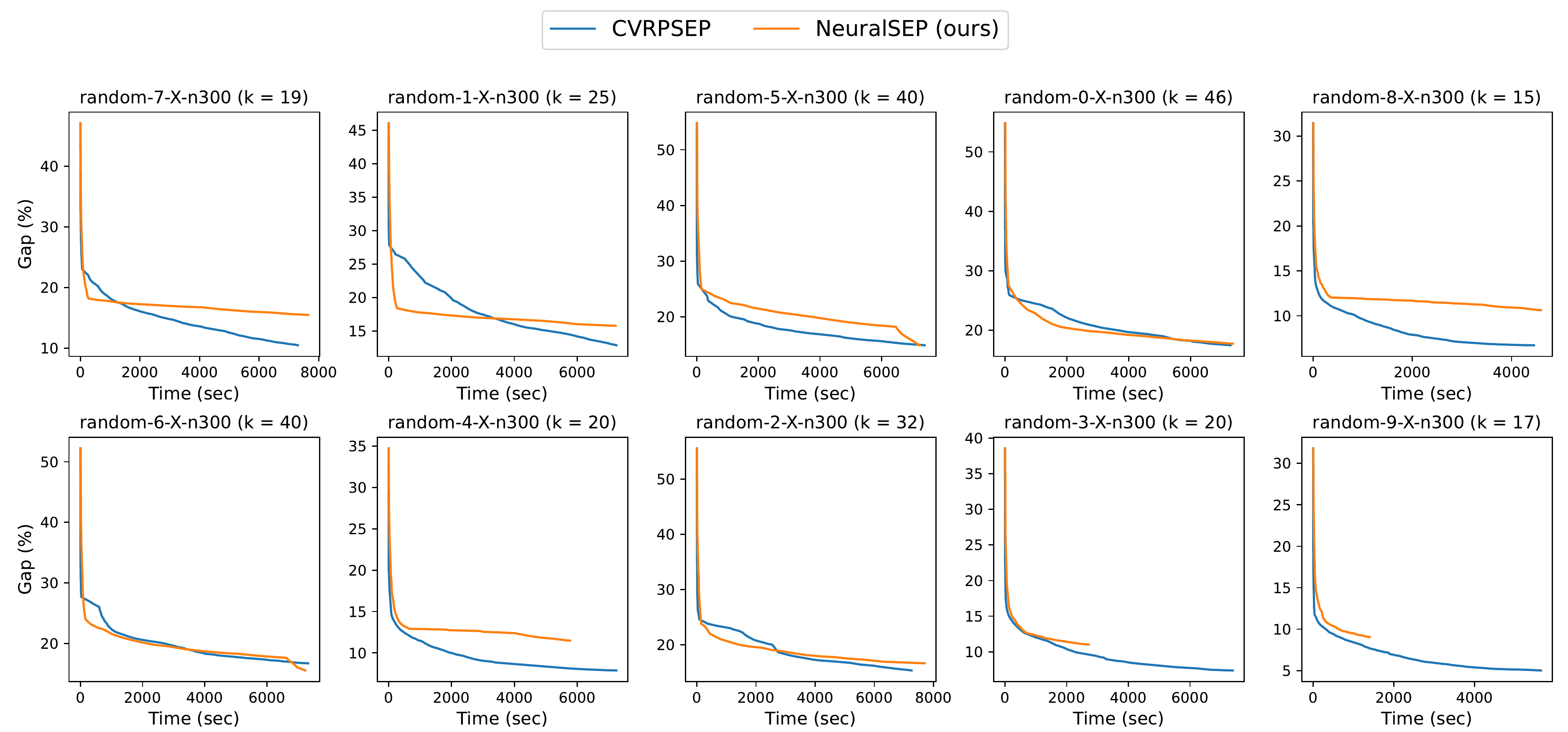}
    \caption{$N = 300$}
    \label{fig:runtime_300}
\end{subfigure}
\hspace*{\fill}
\begin{subfigure}{\textwidth}
    \includegraphics[width=0.9\textwidth]{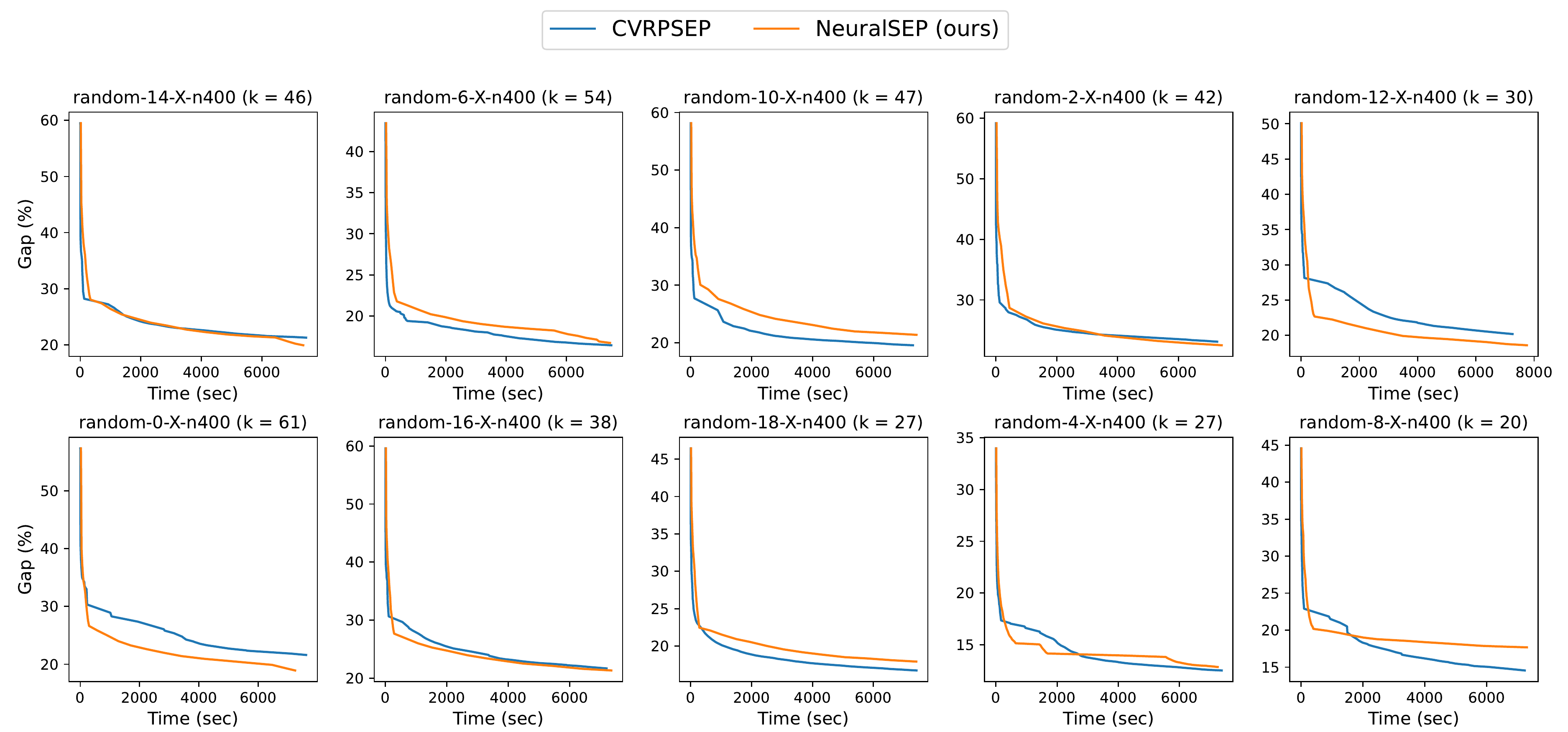}
    \caption{$N = 400$}
    \label{fig:runtime_400}
\end{subfigure}
\caption{Optimality gap comparisons on each instance from $N = 200$ to $N = 400$ with limited time (2h).}
\label{fig:runtime2}
\end{figure}

\begin{figure}
\centering
\begin{subfigure}{\textwidth}
    \includegraphics[width=0.9\textwidth]{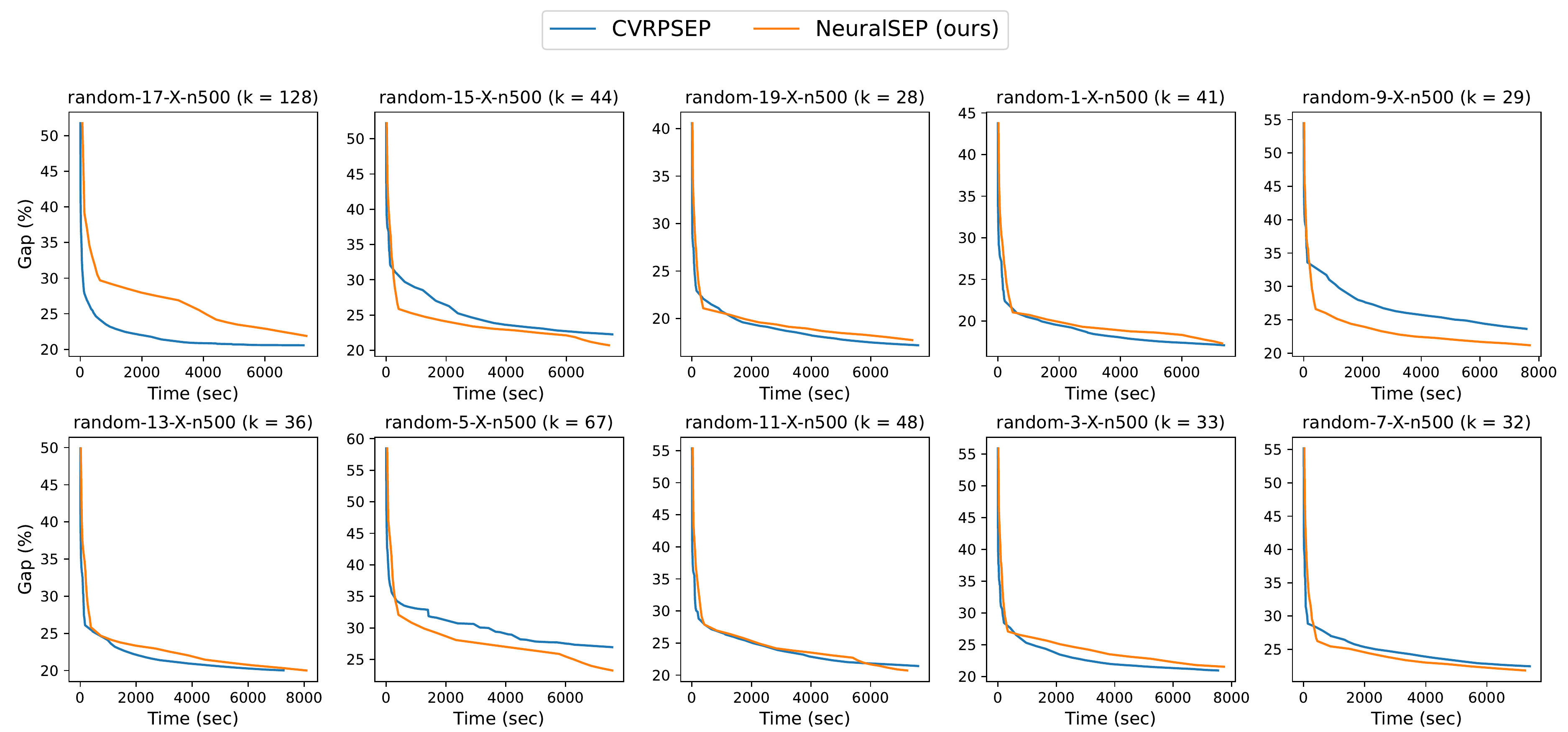}
    \caption{$N = 500$}
    \label{fig:runtime_500}
\end{subfigure}
\hspace*{\fill}
\begin{subfigure}{\textwidth}
    \includegraphics[width=0.9\textwidth]{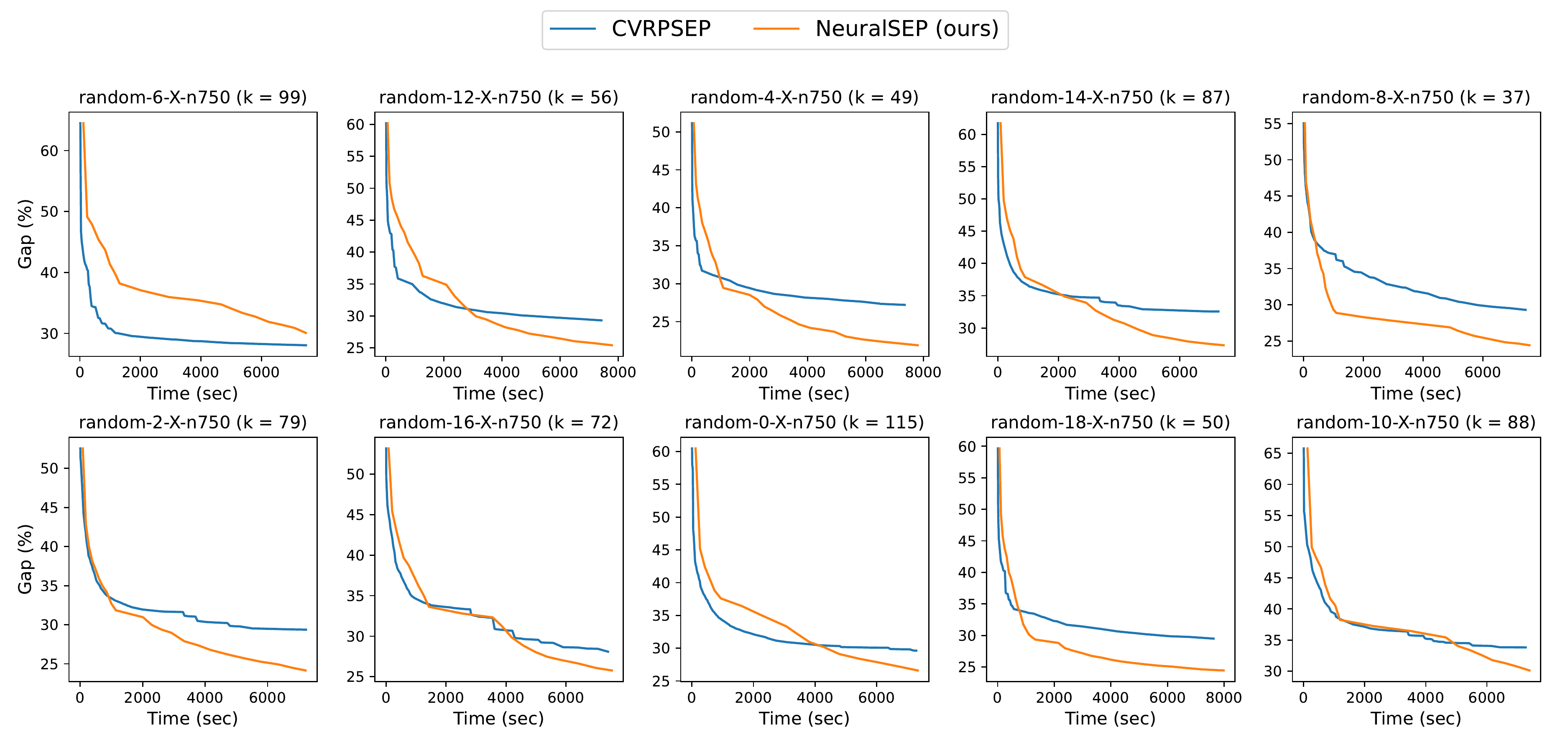}
    \caption{$N = 750$}
    \label{fig:runtime_750}
\end{subfigure}
\hspace*{\fill}
\begin{subfigure}{\textwidth}
    \includegraphics[width=0.9\textwidth]{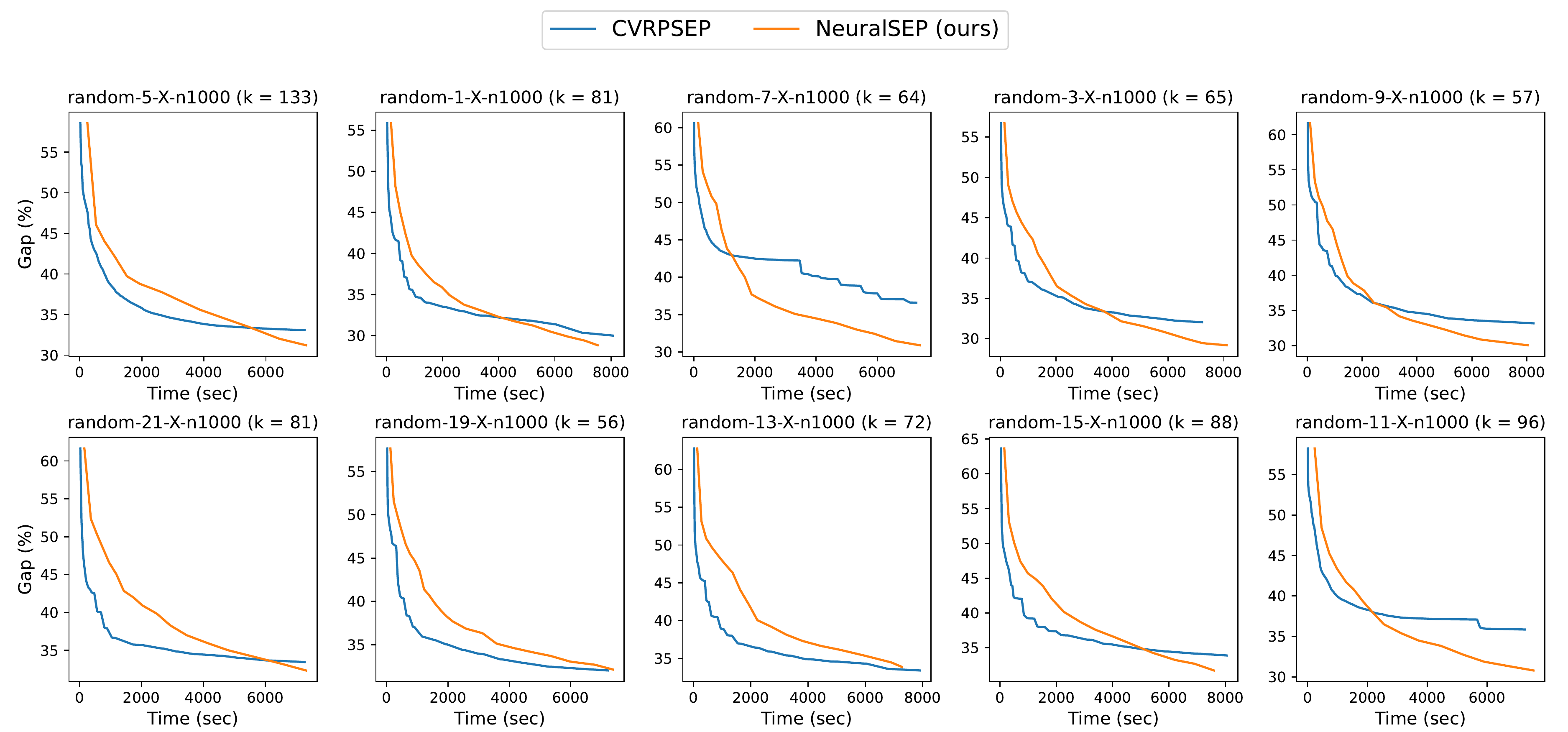}
    \caption{$N = 1000$}
    \label{fig:runtime_1000}
\end{subfigure}
\caption{Optimality gap comparisons on each instance from $N = 500$ to $N = 1000$ with limited time (2h).}
\label{fig:runtime3}
\end{figure}

We provide the optimality gap improvement over time ($\leq$ 2 hours).
Following the guideline of \citet{accorsi2022guidelines}, we compare the performances of CVRPSEP and \ours{} over time, in \cref{fig:runtime1,fig:runtime2,fig:runtime3}.

\newpage
\section{Error Bound of NeuralSEP in Coarsening Procedures}

The experiments measure the average error and ratio of vertices with prediction errors larger than $0.5$ on separation problems, presented in \cref{table:error}.
The test data is collected from the randomly generated CVRP test instances via the exact separation algorithm in advance. 
Since the largest prediction error can exceed $0.5$, NeuralSEP has a possibility of finding infeasible solutions which violate the total demand constraints.
However, cuts are valid for the original CVRP problems by definition regardless of violating these constraints.

\begin{table}[!ht]
\caption{Absolute prediction errors on separation problems.}
\small
\label{table:error}
\centering
\begin{tabular}{rrr} 
\toprule
Size & Avg. Error & Ratio of $\epsilon \geq 0.5$ \\
\midrule
50 & 0.1432 & 0.1401 \\
75 & 0.1337 & 0.1320 \\
100 & 0.1441 & 0.1427 \\ 
200 & 0.1766 & 0.1754 \\
\bottomrule
\end{tabular}
\end{table}

\newpage

\section{Ablation Study} \label{sec:ablation}
We implement two different neuralized separation algorithms for RCIs to verify the effect of graph coarsening. 
The first one is an auto-regressive prediction model like \citet{khalil2017learning} and \citet{park2021schedulenet}, which starts with an empty subset $S$ and sequentially includes a vertex based on the current composition.
The second algorithm is a one-shot prediction model, which directly predicts the selection probability for each vertex at once.
As the probability is a soft assignment, an extra projection scheme, such as rounding \citep{schuetz2022combinatorial}, is required to get a de-randomized graph.

\paragraph{Auto-regressive prediction model.} It takes the support graph and the current partial solution as inputs and decides on which vertices to add to the current solution.
Initially, a dummy vertex is introduced to signify the \emph{`end of selection'}, and edges from other vertices to the dummy are added to aggregate information about the current set composition. 
The edges connected to the dummy vertex have their weights set to 0, and an edge feature is added to indicate that the source vertex presently belongs to the set $S$. 
During training, the model randomly selects a vertex from those labeled as 1. 
The model learns the probability $\hat{p}_i$ in a supervised manner, and the probability is calculated with the exact labels, i.e., 

\begin{equation*}
    \hat{p}_i = \begin{cases}
        \displaystyle\frac{\hat{y}_i}{\sum_{i \in V_c \setminus S} \hat{y}_i} & \quad \mbox{if } i \neq \text{dummy} \\
        1 - \sum_{j \in |V_C|} \hat{p}_j & \quad  \mbox{otherwise.}
    \end{cases}
\end{equation*}
The trained model greedily selects a vertex according to the prediction in the inference phase.

\begin{algorithm}[!htb]
\caption{Training auto-regressive RCI separation} \label{algo:autoregressive}
\small
\begin{algorithmic}[1]
\Require A graph $G$ and label $\hat{y}$
\Ensure Trained parameter $\theta$
    \State Initialize parameter $\theta$
    \State Initialize set $S = \emptyset$
    \For{$v \gets 0$ to $|V_C|-1$}
        \State Compute vertex  probability $\{p_i\}_{i \in V_C \setminus S^{(n)}} \gets f_\theta(G, S)$
        \State Update $\theta \gets \nabla \mathcal{L}(\{p_i\}_{i \in V_C \setminus S}, \{\hat{p}_i\}_{i \in V_C \setminus S})$
        \State Randomly choose a vertex $j$ such that $\hat{y}_i=1$ and $i \in V_C \setminus S$
        \If{$j = \text{dummy}$}
            \State End of selection
        \Else
            \State $S \gets S \cup \{j\}$
        \EndIf
    \EndFor
\end{algorithmic}
\end{algorithm}

\paragraph{One-shot prediction model.}
Using the same input graph with \ours{}, a one-shot prediction model directly calculates the independent vertex probabilities of belonging to $S$. This procedure can be considered as the same as \ours{} without graph coarsening, i.e., a coarsening ratio of $\gamma = 1$. As the model gives the probabilities $\{p_i \}_{i \in V}$ independently, we train the model to imitate the exact label $\hat{y}_i$ as the true selection probability. To discretize the continuous prediction, we employ a simple round rule, which choose vertices with probability higher than 0.5.

\begin{algorithm}[!ht]
\caption{Training one-shot RCI separation}
\label{algo:one-shot}
\small
\begin{algorithmic}[1]
\Require A graph $G$ and label $\hat{y}$
\Ensure Trained parameter $\theta$
    \State Initialize parameter $\theta$
    \State Predict vertex selection probability $\{p_i\}_{i \in V_C} \gets f_\theta(G)$
    \State $p_0 \gets 0$ \Comment{The depot is excluded}
    \State $y_i \gets \text{round}(p_i), \forall i \in V_C$
    \State $S=\{i \in V_C : y_i = 1\}$
    \State Update $\theta \gets \nabla \mathcal{L}(\{p_i\}_{i \in V}, \{\hat{y}_i\}_{i \in V})$
\end{algorithmic}
\end{algorithm}

\begin{figure}
\centering
\hspace*{\fill}
\begin{subfigure}{0.47\textwidth}
    \includegraphics[width=\textwidth]{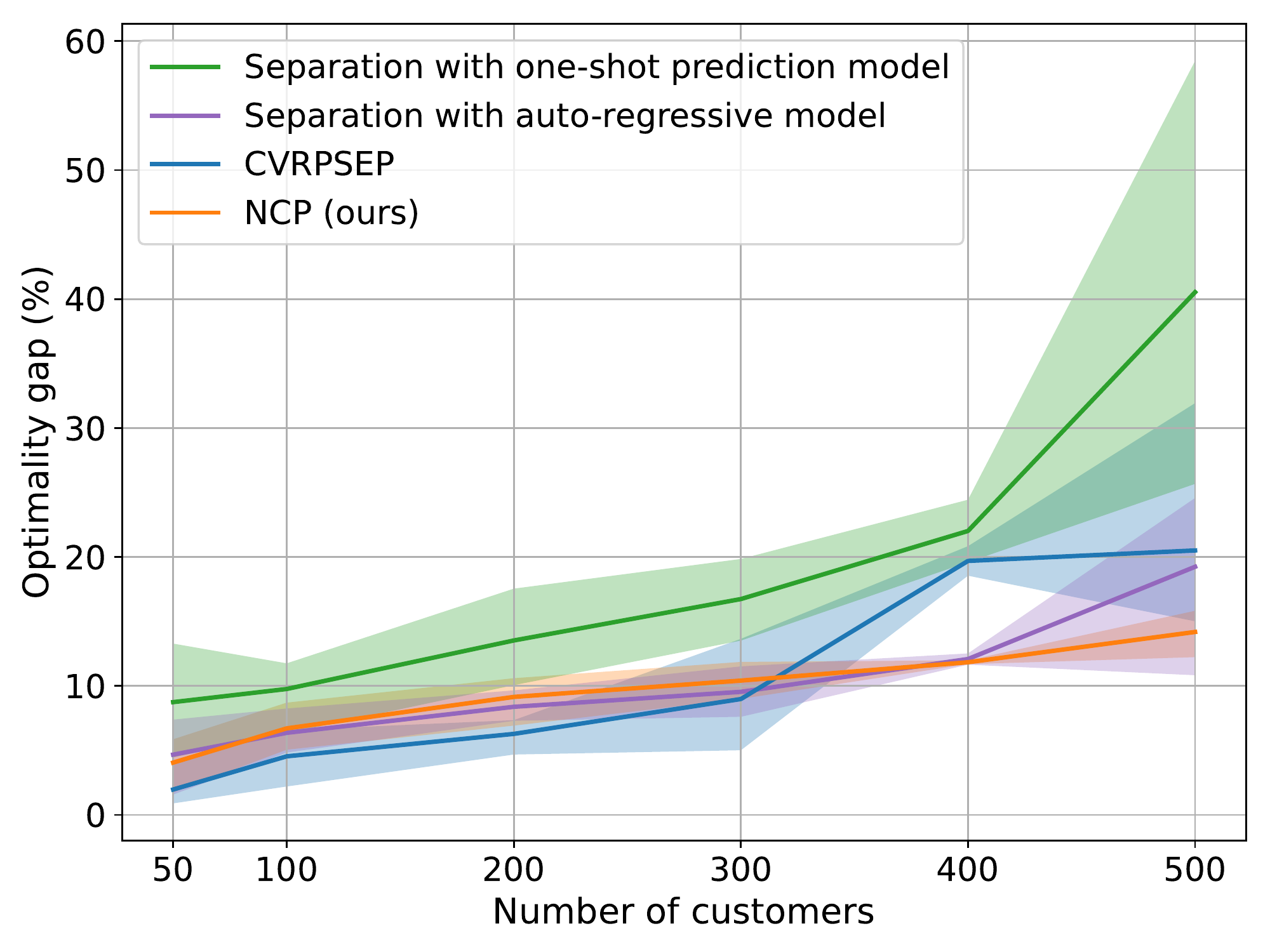}
    \caption{The optimality gap of different neuralized separation algorithms.}
    \label{fig:baseline}
\end{subfigure}
\hspace*{\fill}
\begin{subfigure}{0.47\textwidth}
    \includegraphics[width=\textwidth]{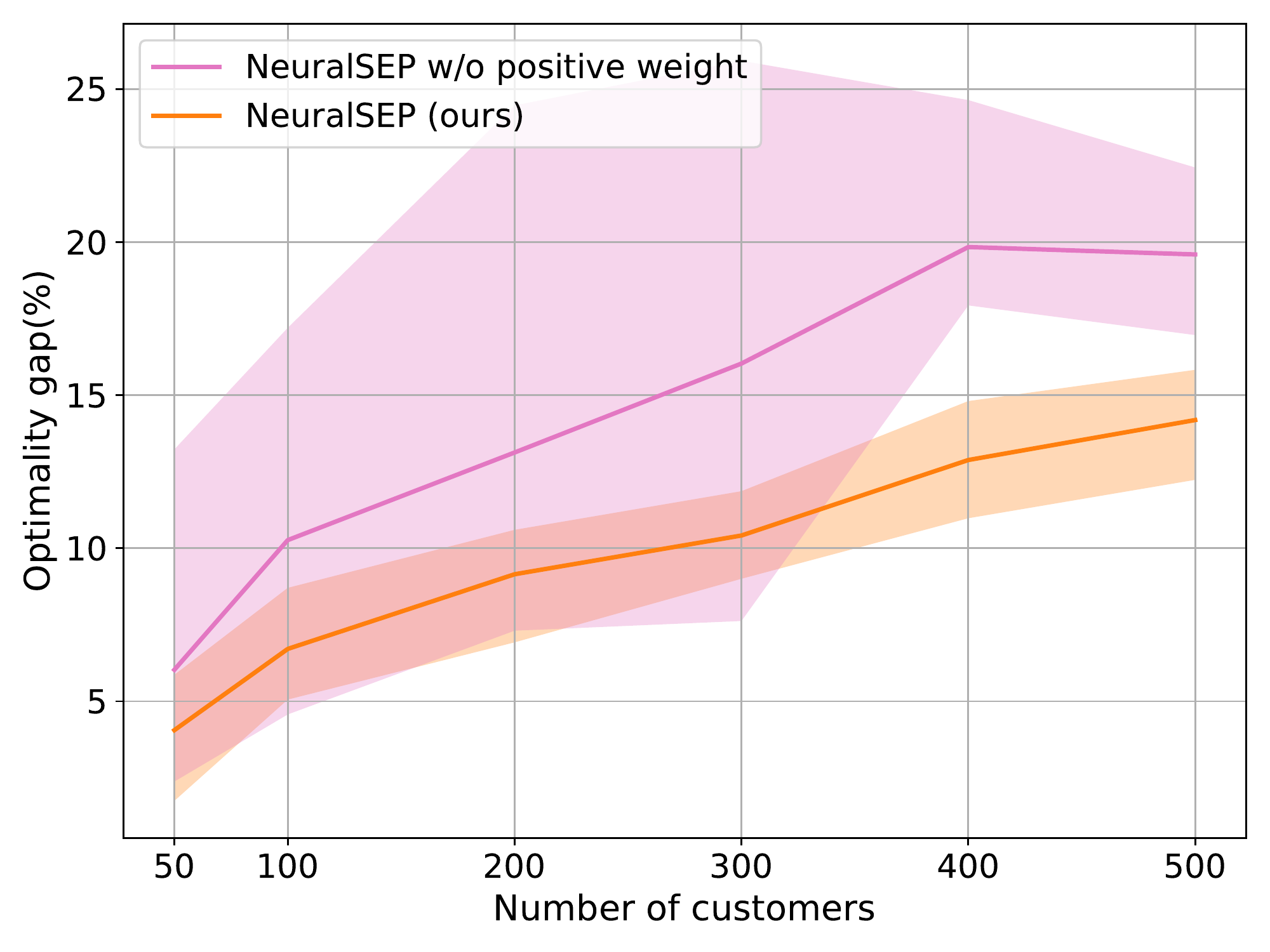}
    \caption{The optimality gap of \ours{} trained with and without the positive weights.}
    \label{fig:ablation}
\end{subfigure}
\caption{{(Ablation) The range of the lower bound gap with HGS.}}
\end{figure}

\paragraph{Comparing performances.}
We compute the lower bound of random CVRP instances in the $[50, 500]$ range. Evaluations are conducted following the same process described in \cref{sec:exp_with_random}.
As shown in \cref{fig:baseline}, our model outperforms the other neuralized models for every test size.
Also, the optimality gap of the one-shot prediction model shows a significant increase when the number of customers exceeds 400. 
We infer that the one-shot model is too challenging to consider the relationship between the vertices (i.e., variables), making it difficult to decide on the vertex assignments jointly.
Further analysis for different neuralized separation algorithms is in \cref{sec:further_exp}.

\begin{table} %
\small
\caption{The resulting lower bounds of cutting plane method with different neuralized separation algorithms.}
\label{table:ablation_details}
\resizebox{\textwidth}{!}{
\centering
\begin{tabular}{l rr rr rr}
	\toprule
	\multicolumn{1}{r}{Size}            & \multicolumn{2}{c}{$50$}  & \multicolumn{2}{c}{$100$} & \multicolumn{2}{c}{$200$} \\ 
										\cmidrule(lr){2-3}          \cmidrule(lr){4-5}          \cmidrule(lr){6-7}	
	\multicolumn{1}{l}{Method}          & Lower bound    &    Iter. & Lower bound    &    Iter. & Lower bound    &    Iter. \\ \midrule
	Auto-regressive                     &     9,083.812  &       41 &     15,447.107 &       90 &     20,519.765 &      169 \\
	One-shot                            &     8,728.685  &       23 &     15,115.604 &       72 &     19,757.546 &      100 \\
	\shortstack{\ours{} w/o pos weight} &     8,922.991  &       26 &     14,847.875 &       48 &     19,593.371 &       94 \\ \midrule
	\textbf{\ours{} (ours)}             & \bf 9,162.498  &       38 & \bf 15,593.334 &       96 & \bf 20,742.544 &      127 \\ \midrule\midrule
	\multicolumn{1}{r}{Size}            & \multicolumn{2}{c}{$300$} & \multicolumn{2}{c}{$400$} & \multicolumn{2}{c}{$500$} \\ 
										\cmidrule(lr){2-3}          \cmidrule(lr){4-5}          \cmidrule(lr){6-7}
	\multicolumn{1}{l}{Method}          & Lower bound    &    Iter. & Lower bound    &    Iter. & Lower bound    &    Iter. \\ \midrule
	Auto-regressive                     &     30,534.670 &      200 &     40,703.404 &      100 &     47,049.873 &      100 \\
	One-shot                            &     28,847.493 &      117 &     38,388.527 &       90 &     36,655.254 &       28 \\
	\shortstack{\ours{} w/o pos weight} &     28,685.375 &       93 &     40,360.798 &       65 &     48,614.153 &       82 \\ \midrule
	\textbf{\ours{} (ours)}             & \bf 31,092.012 &      158 & \bf 43,896.118 &      100 & \bf 53,865.885 &      100 \\ \bottomrule
\end{tabular}}
\end{table}

\paragraph{Effectiveness of the positive weight loss.}
Since the exact labels are highly imbalanced depending on $M$, \ours{} is trained using the positive weighted BCE loss.
To verify the effect of the positive weights, we train the models to minimize the BCE loss with and without the positive weighted loss and compare the optimality gap. \cref{fig:ablation} shows that \ours{} without the positive weights gives a higher average optimality gap for every size.
Furthermore, the optimality gap has larger variances when the model is trained without the positive weights.

\subsection{Performance on Separation Problems} \label{sec:further_exp}
We generate test dataset with the exact separation 
We assess the performance of the models using  the test dataset described in \cref{sec:separation_exp} (100 RCI separation problem instances for each size within the range of $[50, 75, 100, 200]$). We address the RCI separation problem presented in \cref{exact_rci_obj}--\cref{exact_rci_c6} with various neuralized models, including our approach, \ours{}. For each size, we measure the average violations, the average number of required inferences steps, and the ratio of feasible solutions for the separation problem (denoted as Feasibility). As shown in \cref{table:separtion_rst}, clearly indicate that \ours{} requires fewer inference steps to find set $S$, albeit with a slight reduction in constraint satisfaction.

\begin{table} %
\caption{Experimental results on RCI separation with models. The values are reported as averages of 100 separation problem instances.
}
\label{table:separtion_rst}
\small
\centering
\begin{tabular}{lrrrrrr} 
\toprule
\multicolumn{1}{c}{Size} & \multicolumn{3}{c}{$50$}  & \multicolumn{3}{c}{$75$}\\\cmidrule[0.5pt](lr{0.2em}){2-4}\cmidrule[0.5pt](lr{0.2em}){5-7}
 & Violations & Feasibility & Iterations & Violations & Feasibility & Iterations \\
\midrule
Auto-regressive & 1.1126 & \textbf{1.00} & 22.93 & 1.9669 & \textbf{1.00} & 35.09 \\
One-shot & 0.9520 & 0.81 & \textbf{1.00} & 1.7911 & 0.67 & \textbf{1.00}  \\ 
\midrule
\textbf{Coarsening (ours)} & \textbf{1.4579} & 0.65 & 8.89 & \textbf{2.3529} &  0.59 & 9.77 \\
\midrule\midrule
\multicolumn{1}{c}{Size} & \multicolumn{3}{c}{$100$}  & \multicolumn{3}{c}{$200$}\\\cmidrule[0.5pt](lr{0.2em}){2-4}\cmidrule[0.5pt](lr{0.2em}){5-7}
& Violations & Feasibility & Iterations & Violations & Feasibility & Iterations \\
\midrule
Auto-regressive & 1.4095 & \textbf{1.00} & 48.44 & 0.6064 & \textbf{1.00} & 95.86 \\ 
One-shot & 1.2502 & 0.49 & \textbf{1.00} & 0.5090 & 0.63 & \textbf{1.00} \\ 
\midrule
\textbf{Coarsening (ours)} & \textbf{1.8034} & 0.41 & 10.79 & \textbf{0.8764} & 0.37 & 12.96 \\
\bottomrule
\end{tabular}
\end{table}

\subsection{Coarsening with Random Probabilities} 
In this section, we verify the effectiveness of GNN prediction by comparing performances on the separation problems with GNN predictions and the random probabilities. As shown in \cref{table:rand_violation}, \ours{} with GNN succeeds in finding more cuts, and their violations are larger than \ours{} with random probabilities in the average sense. Consequently, \ours{} with GNN tends to have a higher success rate than \ours{} with random probabilities; it shows GNN prediction's fundamental and influential roles in \ours{}.

\label{sec:random_prob_exp}
\begin{table} %
\small
\caption{Performances on separation problems with GNN predictions and random probabilities.}
\label{table:rand_violation}
\centering
\begin{tabular}{rrrrrrr} 
\toprule
\multicolumn{1}{c}{Size} & \multicolumn{2}{c}{Avg. Violations}  & \multicolumn{2}{c}{Num. of Cuts} & \multicolumn{2}{c}{Success Rate} \\ \cmidrule[0.5pt](lr{0.2em}){2-3}\cmidrule[0.5pt](lr{0.2em}){4-5} \cmidrule[0.5pt](lr{0.2em}){6-7}
 & GNN & Random & GNN & Random & GNN & Random \\
\midrule
50 & 2.4540 & 0.6453 & 247 & 79 & 0.68 & 0.33 \\
75 & 4.4779 & 0.3931 & 384 & 214 & 0.56 & 0.34 \\
100 & 3.8272 & 0.3241 & 339 & 92 & 0.49 & 0.28 \\
200 & 3.1650 & 0.1880 & 211 & 66 & 0.34 & 0.14 \\
\bottomrule
\end{tabular}
\end{table}

\clearpage
\section{Experiments with Limited Iterations}
We provide instance-wise results for the experiments with limited iterations on randomly generated CVRP instances and X-instances.

\subsection{CVRPSEP on Randomly Generated Instances with Limited Iterations}

\begin{table}[!ht]
\caption{The results of CVRPSEP on randomly generated instances with limited iterations ($N < 300$).}
\label{table:ec_rand_cvrpsep_small}
\small
\centering
\begin{tabular}{lrrrrrrr}
\toprule
Name & Size & K & Best Known & LB & Gap (\%) & Runtime & Iterations \\
\midrule
random-24-X-n50 & 50 & 3 & 6,686 & 6,542.30 & 2.15 & 22.42 & 21  \\
random-4-X-n50 & 50 & 4 & 7,684 & 7,579.00 & 1.37 & 0.13 & 20  \\
random-28-X-n50 & 50 & 7 & 11,557 & 11,081.25 & 4.12 & 0.50 & 35  \\
random-0-X-n50 & 50 & 8 & 12,735 & 12,237.23 & 3.91 & 0.38 & 28  \\
random-16-X-n50 & 50 & 5 & 10,190 & 10,091.62 & 0.97 & 0.33 & 28  \\
random-20-X-n50 & 50 & 6 & 10,126 & 9,929.22 & 1.94 & 0.25 & 22  \\
random-32-X-n50 & 50 & 4 & 7,483 & 7,395.72 & 1.17 & 0.15 & 19  \\
\bottomrule
\end{tabular}

\end{table}

\begin{table}[!ht]
\caption{The results of CVRPSEP on randomly generated instances with limited iterations ($N \geq 300$).}
\label{table:ec_rand_cvrpsep_large}
\renewcommand\arraystretch{0.9}
\small
\centering
\begin{tabular}{lrrrrrrr}
\toprule
Name & Size & K & Best Known & LB & Gap (\%) & Runtime & Iteration \\
\midrule
random-7-X-n300 & 300 & 19 & 27,269 & 25,089.83 & 7.99 & 10,752.32 & 200  \\
random-1-X-n300 & 300 & 25 & 30,181 & 27,441.55 & 9.08 & 7,393.51 & 200  \\
random-5-X-n300 & 300 & 40 & 53,305 & 48,350.51 & 9.29 & 11,862.99 & 200  \\
random-0-X-n300 & 300 & 46 & 55,614 & 48,022.52 & 13.65 & 9,509.19 & 200  \\
random-8-X-n300 & 300 & 15 & 20,636 & 19,246.52 & 6.73 & 1,804.49 & 119  \\
random-6-X-n300 & 300 & 40 & 42,079 & 37,349.29 & 11.24 & 13,628.17 & 200  \\
random-4-X-n300 & 300 & 20 & 24,324 & 22,420.79 & 7.82 & 3,313.94 & 150  \\
\bottomrule
\end{tabular}

\end{table}

\clearpage
\subsection{NeuralSEP on Randomly Generated Instances with Limited Iterations}

\begin{table}[!ht]
\caption{The results of \ours{} on randomly generated instances with limited iterations ($N < 300$).}
\label{table:ec_rand_neuralsep_small}
\small
\centering
\begin{tabular}{lrrrrrrr}
\toprule
Name & Size & K & Best Known & LB & Gap (\%) & Runtime & Iterations \\
\midrule
random-24-X-n50 & 50  & 3  & 6,686  & 6,367.50  & 4.76  & 44.32  & 24  \\
random-4-X-n50 & 50  & 4  & 7,684  & 7,371.75  & 4.06  & 44.03  & 30  \\
random-28-X-n50 & 50  & 7  & 11,557  & 10,879.39  & 5.86  & 85.94  & 44  \\
random-0-X-n50 & 50  & 8  & 12,735  & 11,988.86  & 5.86  & 77.72  & 46  \\
random-16-X-n50 & 50  & 5  & 10,190  & 9,861.49  & 3.22  & 67.56  & 45  \\
random-20-X-n50 & 50  & 6  & 10,126  & 9,673.80  & 4.47  & 61.03  & 38  \\
random-32-X-n50 & 50  & 4  & 7,483  & 7,271.80  & 2.82  & 46.29  & 34  \\
\bottomrule
\end{tabular}
\end{table}

\begin{table}[!ht]
\caption{The results of \ours{} on randomly generated instances with limited iterations ($N \geq 300$).}
\label{table:ec_rand_neuralsep_large}
\renewcommand\arraystretch{0.9}
\small
\centering
\begin{tabular}{lrrrrrrr}
\toprule
Name & Size & K & Best Known & LB & Gap (\%) & Runtime & Iterations \\
\midrule
random-7-X-n300 & 300  & 19  & 27,269  & 24,034.53  & 11.86  & 1,862.09  & 141   \\
random-1-X-n300 & 300  & 25  & 30,181  & 26,725.37  & 11.45  & 2,885.56  & 163   \\
random-5-X-n300 & 300  & 40  & 53,305  & 48,507.47  & 9.00  & 11,676.94  & 200   \\
random-0-X-n300 & 300  & 46  & 55,614  & 50,157.63  & 9.81  & 16,801.28  & 200   \\
random-8-X-n300 & 300  & 15  & 20,636  & 18,437.43  & 10.65  & 862.25  & 106   \\
random-6-X-n300 & 300  & 40  & 42,079  & 37,803.43  & 10.16  & 9,308.73  & 198   \\
random-4-X-n300 & 300  & 20  & 24,324  & 21,475.71  & 11.71  & 1,097.75  & 108   \\
\bottomrule
\end{tabular}
\end{table}

\clearpage
\subsection{CVRPSEP on X-instances with Limited Iterations}

\begin{table}[!ht]
\caption{The results of CVRPSEP on X-instances with limited iterations ($N < 300$).}
\label{table:ec_x_cvrpsep_small}
\small
\centering
\begin{tabular}{lrrrrrrr}
\toprule
Name & Size & K & Best Known & LB & Gap (\%) & Runtime & Iteration \\
\midrule
X-n101-k25 & 100 & 25 & 27,591.00 & 26,512.38 & 3.91 & 28.97 & 61 \\
X-n106-k14 & 105 & 14 & 26,362.00 & 25,879.17 & 1.83 & 64.33 & 106 \\
X-n110-k13 & 109 & 13 & 14,971.00 & 14,308.15 & 4.43 & 8.83 & 37 \\
X-n115-k10 & 114 & 10 & 12,747.00 & 12,359.04 & 3.04 & 9.28 & 38 \\
X-n120-k6 & 119 & 6 & 13,332.00 & 12,681.53 & 4.88 & 14.21 & 48 \\
X-n125-k30 & 124 & 30 & 55,539.00 & 53,923.10 & 2.91 & 143.55 & 133 \\
X-n129-k18 & 128 & 18 & 28,940.00 & 27,568.78 & 4.74 & 135.38 & 121 \\
\bottomrule
\end{tabular}

\end{table}

\begin{table}[!ht]
\caption{The results of CVRPSEP on X-instances with limited iterations ($N \geq 300$).}
\label{table:ec_x_cvrpsep_large}
\renewcommand\arraystretch{0.9}
\small
\centering
\begin{tabular}{lrrrrrrr}
\toprule
Name & Size & K & Best Known & LB & Gap (\%) & Runtime & Iteration \\
\midrule
X-n303-k21 &  302  &  21  &  21,736.00  &  20,206.01  &  7.04  &  1,853.02  &  100 \\
X-n308-k13 &  307  &  13  &  25,859.00  &  23,604.19  &  8.72  &  2,509.98  &  100 \\
X-n313-k71 &  312  &  71  &  94,043.00  &  80,064.40  &  14.86  &  639.29  &  100 \\
X-n317-k53 &  316  &  53  &  78,355.00  &  62,504.84  &  20.23  &  924.83  &  100 \\
X-n322-k28 &  321  &  28  &  29,834.00  &  26,567.21  &  10.95  &  2,910.12  &  100 \\
X-n327-k20 &  326  &  20  &  27,532.00  &  23,951.60  &  13.00  &  2,471.64  &  100 \\
X-n331-k15 &  330  &  15  &  31,102.00  &  28,411.45  &  8.65  &  3,660.40  &  100 \\
\bottomrule
\end{tabular}

\end{table}

\clearpage
\subsection{NeuralSEP on X-instances with Limited Iterations}

\begin{table}[!ht]
\caption{The results of \ours{} on X-instances with limited iterations ($N < 300$).}
\label{table:ec_x_neuralsep_small}
\small
\centering
\begin{tabular}{lrrrrrrr}
\toprule
Name & Size & K & Best Known & LB & Gap (\%) & Runtime & Iteration \\
\midrule
X-n101-k25 & 100 & 25 &  27,591  &  25,910.93  &  6.09  &  403.91  &  146 \\
X-n106-k14 & 105 & 14 &  26,362  &  25,571.17  &  3.00  &  259.14  &  116 \\
X-n110-k13 & 109 & 13 &  14,971  &  14,053.06  &  6.13  &  201.96  &  100 \\
X-n115-k10 & 114 & 10 &  12,747  &  11,872.71  &  6.86  &  84.15  &  44 \\
X-n120-k6 & 119 & 6 &  13,332  &  12,259.17  &  8.05  &  120.19  &  66 \\
X-n125-k30 & 124 & 30 &  55,539  &  53,337.66  &  3.96  &  696.05  &  164 \\
X-n129-k18 & 128 & 18 &  28,940  &  27,302.55  &  5.66  &  572.17  &  169 \\
\bottomrule
\end{tabular}

\end{table}

\begin{table}[!ht]
\caption{The results of \ours{} on X-instances with limited iterations ($N \geq 300$).}
\label{table:ec_x_neuralsep_large}
\renewcommand\arraystretch{0.9}
\small
\centering
\begin{tabular}{lrrrrrrr}
\toprule
Name & Size & K & Best Known & LB & Gap (\%) & Runtime & Iteration \\
\midrule
X-n303-k21 & 302 & 21 &  21,736  &  19,812.05  &  8.85  &  1,169.65  &  100 \\
X-n308-k13 & 307 & 13 &  25,859  &  22,805.38  &  11.81  &  924.02  &  100 \\
X-n313-k71 & 312 & 71 &  94,043  &  84,502.88  &  10.14  &  16,158.87  &  100 \\
X-n317-k53 & 316 & 53 &  78,355  &  70,831.31  &  9.60  &  7,938.16  &  100 \\
X-n322-k28 & 321 & 28 &  29,834  &  26,332.21  &  11.74  &  1,931.83  &  100 \\
X-n327-k20 & 326 & 20 &  27,532  &  23,505.43  &  14.63  &  1,589.05  &  100 \\
X-n331-k15 & 330 & 15 &  31,102  &  27,322.48  &  12.15  &  1,079.60  &  100 \\
\bottomrule
\end{tabular}

\end{table}

\clearpage
\section{Experiments with Limited Time}
We provide instance-wise results for the experiments with limited time (2 hours) on randomly generated CVRP instances and X-instances.

\subsection{CVRPSEP on Randomly Generated Instances with Limited Time}

\begin{table}[!ht]
\caption{The results of CVRPSEP on randomly generated instances with 2 hours limit ($N < 300$).}
\small
\centering
\begin{tabular}{lrrrrrrr}
\toprule
Name & Size & K & Best Known & LB & Gap (\%) & Runtime & Iterations \\
\midrule
random-24-X-n50 & 50 & 3 & 6,686.00 & 6,542.30 & 2.15 & 0.13 & 24 \\
random-4-X-n50 & 50 & 4 & 7,684.00 & 7,579.00 & 1.37 & 0.15 & 23 \\
random-28-X-n50 & 50 & 7 & 11,557.00 & 11,081.25 & 4.12 & 0.66 & 37 \\
random-0-X-n50 & 50 & 8 & 12,735.00 & 12,237.15 & 3.91 & 0.55 & 30 \\
random-16-X-n50 & 50 & 5 & 10,190.00 & 10,091.62 & 0.97 & 0.42 & 30 \\
random-20-X-n50 & 50 & 6 & 10,126.00 & 9,929.22 & 1.94 & 0.28 & 21 \\
random-32-X-n50 & 50 & 4 & 7,483.00 & 7,395.72 & 1.17 & 0.20 & 21 \\
\bottomrule
\end{tabular}
\end{table}

\begin{table}[!ht]
\caption{The results of CVRPSEP on randomly generated instances with 2 hours limit ($N \geq 300$).}
\small
\centering
\begin{tabular}{lrrrrrrr}
\toprule
Name & Size & K & Best Known & LB & Gap (\%) & Runtime & Iterations \\
\midrule
random-7-X-n300 & 300 & 19 & 27,269.00 & 24,406.82 & 10.50 & 7,298.81 & 101 \\
random-1-X-n300 & 300 & 25 & 30,181.00 & 26,291.65 & 12.89 & 7,245.98 & 103 \\
random-5-X-n300 & 300 & 40 & 53,305.00 & 45,354.34 & 14.92 & 7,394.68 & 96 \\
random-0-X-n300 & 300 & 46 & 55,614.00 & 45,887.00 & 17.49 & 7,293.01 & 139 \\
random-8-X-n300 & 300 & 15 & 20,636.00 & 19,247.28 & 6.73 & 4,451.55 & 121 \\
random-6-X-n300 & 300 & 40 & 42,079.00 & 35,040.64 & 16.73 & 7,326.92 & 100 \\
random-4-X-n300 & 300 & 20 & 24,324.00 & 22,408.96 & 7.87 & 7,240.73 & 133 \\
\bottomrule
\end{tabular}
\end{table}

\newpage
\subsection{NeuralSEP on Randomly Generated Instances with Limited Time}
\begin{table}[!ht]
\caption{The results of NeuralSEP on randomly generated instances with 2 hours limit ($N < 300$).}
\small
\centering
\begin{tabular}{lrrrrrrr}
\toprule
Name & Size & K & Best Known & LB & Gap (\%) & Runtime & Iterations \\
\midrule
random-24-X-n50 & 50 & 3 &  6,686.00  &  6,367.50  &  4.76  &  31.68  &  25 \\
random-4-X-n50 & 50 & 4 &  7,684.00  &  7,349.22  &  4.36  &  24.37  &  32 \\
random-28-X-n50 & 50 & 7 &  11,557.00  &  10,941.28  &  5.33  &  47.55  &  59 \\
random-0-X-n50 & 50 & 8 &  12,735.00  &  12,047.69  &  5.40  &  47.37  &  58 \\
random-16-X-n50 & 50 & 5 &  10,190.00  &  9,888.76  &  2.96  &  32.30  &  47 \\
random-20-X-n50 & 50 & 6 &  10,126.00  &  9,615.55  &  5.04  &  19.20  &  32 \\
random-32-X-n50 & 50 & 4 &  7,483.00  &  7,202.71  &  3.75  &  23.01  &  35 \\
\bottomrule
\end{tabular}
\end{table}

\begin{table}[!ht]
\caption{The results of NeuralSEP on randomly generated instances with 2 hours limit ($N \geq 300$).}
\small
\centering
\begin{tabular}{lrrrrrrr}
\toprule
Name & Size & K & Best Known & LB & Gap (\%) & Runtime & Iterations \\
\midrule
random-7-X-n300 & 300 & 19 &  27,269.00  &  23,044.48  &  15.49  &  7,644.58  &  40 \\
random-1-X-n300 & 300 & 25 &  30,181.00  &  25,414.03  &  15.79  &  7,221.22  &  34 \\
random-5-X-n300 & 300 & 40 &  53,305.00  &  45,360.13  &  14.90  &  7,243.09  &  43 \\
random-0-X-n300 & 300 & 46 &  55,614.00  &  45,744.42  &  17.75  &  7,363.66  &  27 \\
random-8-X-n300 & 300 & 15 &  20,636.00  &  18,441.22  &  10.64  &  4,591.89  &  106 \\
random-6-X-n300 & 300 & 40 &  42,079.00  &  35,537.00  &  15.55  &  7,232.09  &  41 \\
random-4-X-n300 & 300 & 20 &  24,324.00  &  21,529.87  &  11.49  &  5,766.47  &  120 \\
\bottomrule
\end{tabular}
\end{table}

\newpage
\subsection{CVRPSEP on X-Instances with Limited Time}

\begin{table}[!ht]
\caption{The results of CVRPSEP on X-instances with 2 hours limit ($N < 300$).}
\small
\centering
\begin{tabular}{lrrrrrrr}
\toprule
Name & Size & K & Best Known & LB & Gap (\%) & Runtime & Iterations \\
\midrule
X-n101-k25 & 100 & 25 & 27,591.00 & 26,515.75 & 3.90 & 30.18 & 61 \\
X-n106-k14 & 105 & 14 & 26,362.00 & 25,879.30 & 1.83 & 77.18 & 107 \\
X-n110-k13 & 109 & 13 & 14,971.00 & 14,308.37 & 4.43 & 10.95 & 40 \\
X-n115-k10 & 114 & 10 & 12,747.00 & 12,359.39 & 3.04 & 13.12 & 41 \\
X-n120-k6 & 119 & 6 & 13,332.00 & 12,681.59 & 4.88 & 17.64 & 47 \\
X-n125-k30 & 124 & 30 & 55,539.00 & 53,924.08 & 2.91 & 181.01 & 137 \\
X-n129-k18 & 128 & 18 & 28,940.00 & 27,567.53 & 4.74 & 223.4 & 120 \\
\bottomrule
\end{tabular}
\end{table}

\begin{table}[!ht]
\caption{The results of CVRPSEP on X-instances with 2 hours limit ($N \geq 300$).}
\renewcommand\arraystretch{0.9}
\small
\centering
\begin{tabular}{lrrrrrrr}
\toprule
Name & Size & K & Best Known & LB & Gap (\%) & Runtime & Iterations \\
\midrule
X-n303-k21 & 302 & 21 & 21,736.00 & 20,337.11 & 6.44 & 7,321.90 & 123 \\
X-n308-k13 & 307 & 13 & 25,859.00 & 23,567.35 & 8.86 & 7,239.33 & 88 \\
X-n313-k71 & 312 & 71 & 94,043.00 & 81,237.09 & 13.62 & 7,206.92 & 100 \\
X-n317-k53 & 316 & 53 & 78,355.00 & 65,549.59 & 16.34 & 7,360.92 & 98 \\
X-n322-k28 & 321 & 28 & 29,834.00 & 26,218.74 & 12.12 & 7,204.95 & 76 \\
X-n327-k20 & 326 & 20 & 27,532.00 & 23,880.89 & 13.26 & 7,362.26 & 108 \\
X-n331-k15 & 330 & 15 & 31,102.00 & 27,794.10 & 10.64 & 7,361.60 & 76 \\
\bottomrule
\end{tabular}
\end{table}

\newpage
\subsection{NeuralSEP on X-Instances with Limited Time}

\begin{table}[!ht]
\caption{The results of NeuralSEP on X-instances with 2 hours limit ($N < 300$).}
\small
\centering
\begin{tabular}{lrrrrrrr}
\toprule
Name & Size & K & Best Known & LB & Gap (\%) & Runtime & Iterations \\
\midrule
X-n101-k25 & 100 & 25 &  27,591.00  &  25,918.66  &  6.06  &  389.39  &  144 \\
X-n106-k14 & 105 & 14 &  26,362.00  &  25,616.21  &  2.83  &  332.18  &  145 \\
X-n110-k13 & 109 & 13 &  14,971.00  &  14,001.96  &  6.47  &  153.15  &  89 \\
X-n115-k10 & 114 & 10 &  12,747.00  &  12,000.24  &  5.86  &  90.94  &  55 \\
X-n120-k6 & 119 & 6 &  13,332.00  &  12,243.66  &  8.16  &  80.28  &  49 \\
X-n125-k30 & 124 & 30 &  55,539.00  &  53,400.40  &  3.85  &  1,015.84  &  201 \\
X-n129-k18 & 128 & 18 &  28,940.00  &  27,238.26  &  5.88  &  789.12  &  187 \\
\bottomrule
\end{tabular}
\end{table}

\begin{table}[!ht]
\caption{The results of NeuralSEP on X-instances with 2 hours limit ($N \geq 300$).}
\renewcommand\arraystretch{0.9}
\small
\centering
\begin{tabular}{lrrrrrrr}
\toprule
Name & Size & K & Best Known & LB & Gap (\%) & Runtime & Iterations \\
\midrule
X-n303-k21 & 302 & 21 &  21,736.00  &  19,938.02  &  8.27  &  3,933.31  &  154 \\
X-n308-k13 & 307 & 13 &  25,859.00  &  22,397.40  &  13.39  &  7,274.41  &  46 \\
X-n313-k71 & 312 & 71 &  94,043.00  &  75,117.74  &  20.12  &  7,263.09  &  24 \\
X-n317-k53 & 316 & 53 &  78,355.00  &  62,199.48  &  20.62  &  7,411.41  &  28 \\
X-n322-k28 & 321 & 28 &  29,834.00  &  26,488.19  &  11.21  &  7,202.03  &  167 \\
X-n327-k20 & 326 & 20 &  27,532.00  &  22,688.84  &  17.59  &  7,604.80  &  35 \\
X-n331-k15 & 330 & 15 &  31,102.00  &  26,659.59  &  14.28  &  7,806.68  &  46 \\
\bottomrule
\end{tabular}
\end{table}

\end{document}